\documentclass[accepted]{uai2021} % after acceptance, for a revised
\usepackage[american]{babel}
\usepackage{natbib} % has a nice set of citation styles and commands
\usepackage{mathtools} % amsmath with fixes and additions
\usepackage{booktabs} % commands to create good-looking tables
\usepackage{graphicx}
\usepackage{multirow}
\usepackage[table]{xcolor}
\usepackage{bm}
\usepackage{amssymb}
\usepackage{xr}
\newcommand*\samethanks[1][\value{footnote}]{\footnotemark[#1]}

\usepackage[ruled,linesnumbered,noend]{algorithm2e}
\newcommand{\var}{\texttt}
\SetKwComment{Comment}{$\triangleleft$\ }{}

\SetCommentSty{mycommfont}

\usepackage{amsthm}
\newtheorem{theorem}{Theorem}[section]
\theoremstyle{definition}
\newtheorem{definition}{Definition}[section]

\newtheorem{lemma}{Lemma}[section]

\usepackage{array,environ}
\DeclareMathOperator*{\minimize}{\text{minimize}}

\newcolumntype{R}{>{\displaystyle}r}
\newcolumntype{L}{>{\displaystyle}l}
\newenvironment{problem}
  {%
   \begin{array}{R@{\quad}L}}
  {\end{array}}

\usepackage{appendix}
\usepackage{soul}

\usepackage{pifont}% http://ctan.org/pkg/pifont
\newcommand{\cmark}{\ding{52}}%
\newcommand{\xmark}{\ding{56}}%

\title{Improving Uncertainty Calibration of Deep Neural Networks via Truth Discovery and Geometric Optimization}

\author[1]{\href{mailto:Chunwei Ma <chunweim@buffalo.edu>?Subject=Your UAI 2021 paper}{Chunwei Ma}{}}
\author[2]{\href{mailto:Ziyun Huang <zxh201@psu.edu>?Subject=Your UAI 2021 paper}{Ziyun Huang}{}}
\author[1]{\href{mailto:Jiayi Xian <jxian@buffalo.edu>?Subject=Your UAI 2021 paper}{Jiayi Xian}{}}
\author[1]{\href{mailto:Mingchen Gao <mgao8@buffalo.edu>?Subject=Your UAI 2021 paper}{Mingchen Gao\thanks{Co-corresponding authors.}}{}}
\author[1]{\href{mailto:Jinhui Xu <jinhui@buffalo.edu>?Subject=Your UAI 2021 paper}{Jinhui Xu\samethanks{}}{}}

\affil[1]{%
    Department of Computer Science and Engineering\\
    University at Buffalo\\
    Buffalo, NY, USA
}
\affil[2]{%
    Computer Science and Software Engineering\\
    Penn State Erie\\
    Erie, PA, USA
}

\begin{document}
\maketitle
\renewcommand{\thefootnote}{\fnsymbol{footnote}}

% ==================== ==================== ==================== ==================== ====================
% 0 Abstract
% ==================== ==================== ==================== ==================== ====================
\begin{abstract}
  Deep Neural Networks (DNNs), despite their tremendous success in recent years, could still cast doubts on their predictions due to the intrinsic uncertainty associated with their learning process.
  %with their extraordinary performance, could be more trusted if the uncertainty of their predictions can be quantified. 
  Ensemble techniques and post-hoc calibrations are two types of approaches that have individually shown promise in  %significantly 
  improving 
  %can significantly improve 
  the uncertainty calibration of DNNs. %-----A major remaining question 
  %facing the research community 
                                       %-----is thus to determine whether they can be
  %that they are usually performed separately, and it is not clear whether they can be 
                                       %-----used together to complement each other.
  %however, the two are usually performed separately. 
  %~~~~~revised on 2-20~~~~~%
  However, the synergistic effect of the two types of methods has not been well explored.
  In this paper, we propose a truth discovery framework to integrate ensemble-based and post-hoc calibration methods.  %under a truth discovery framework. 
  %Our framework uses 
  %Our framework uses 
  Using 
  the geometric variance of the ensemble candidates as a good indicator for sample uncertainty, 
  %and allows us to obtain 
  %our framework enables us to obtain 
  we design 
  %show that an 
  an accuracy-preserving 
  truth estimator with provably no 
  %preserves  has provably no 
  accuracy drop. Furthermore, 
  %our framework can also enhance 
  we show that 
  post-hoc calibration can also be 
  % promoted 
  enhanced by truth discovery-regularized optimization. 
  %equipped with a truth discovery regularizer.  %optimization.
                                       %-----We validated our methods on large-scale datasets including CIFAR and ImageNet, with both histogram-based and kernel density-based evaluation metrics.
                                       %-----To our best knowledge, this is the first result to bridge the divide between ensemble-based and post-hoc calibration methods.  
  %~~~~~revised on 2-20~~~~~% may not be the fist
  On large-scale datasets including CIFAR and ImageNet, our method shows consistent improvement against state-of-the-art calibration approaches on both histogram-based and kernel density-based evaluation metrics.
  Our code is available at https://github.com/horsepurve/truly-uncertain.
\end{abstract}

% ==================== ==================== ==================== ==================== ====================
% 1 Introduction
% ==================== ==================== ==================== ==================== ====================
\section{Introduction}\label{sec:intro}

We live in an uncertain world. With the increasing use of deep learning in the real world, quantitative estimation of the predictions from deep neural networks (DNNs) must not be neglected, especially when it comes to medical imaging \citep{esteva2017dermatologist} \citep{ma2019neural}, disease diagnosis \citep{de2018clinically} \citep{ma2018improved}, and autonomous driving \citep{kendall2017end}. Uncertainty also plays an important role in differentially private data analysis  \citep{FOCSS13}. 

Modern deep neural networks, despite their extraordinary performance, are oft-criticized as being poorly calibrated and prone to be overconfident, thus leading to unsatisfied uncertainty estimation. The process of adapting deep learning's output to be consistent with the actual probability is called \emph{uncertainty calibration} \citep{guo2017calibration}, and has drawn a growing attention in recent years.

% ========== uncertainty research review & their limitations & current challenges ==========
%Past efforts for achieving effective uncertainty calibration of DNNs have been concentrated on
For a better calibration of the uncertainty of DNNs, the efforts to date have been 
%devoted to 
%the efforts to date have been 
concentrated on
%devoted to 
developing more effective calibration and evaluation methods. 
%and designing better evaluation methods. 
%Modern 
Existing calibration methods roughly fall into two categories, depending on whether 
%in terms of if 
an additional hold-out calibration dataset is used. (1) Post-hoc calibration methods use a calibration dataset to
learn a parameterized transformation that maps from classifiers' raw outputs to their expected probabilities. 
%using the calibration dataset. 
%There are 
Quite a few techniques in this category can be used to learn the mapping, such as 
%The mapping can be learned by 
Temperature Scaling (TS) \citep{guo2017calibration} \citep{kull2019beyond}, Ensemble Temperature Scaling (ETS) \citep{mixnmatch}, and cubic spline \citep{gupta2021calibration}, etc.  However, the expressivity of the learnable mapping could still be limited in all of them. This is evidenced by the fact that 
%For example, 
in TS a single temperature parameter $T$ is tuned, while ETS brings in three additional ensemble parameters. Thus, it is desirable to explore a more sophisticated form of the mapping function. 
%Therefore, the possibility of a more sophisticated form of this mapping function has not been well explored. 
(2) Another line of methods adapt the training process so that the predictions are better calibrated. Techniques in this category include %which is accomplished by 
mixup training \citep{thulasidasan2019mixup}, pre-training \citep{hendrycks2019pre}, label-smoothing \citep{NEURIPS2019_When}, data augmentation \citep{Ashukha2020Pitfalls}, self-supervised learning \citep{hendrycks2019self}, Bayesian approximation \citep{gal2016dropout} \citep{Gal2017ConcreteB}, and Deep Ensemble (DE) \citep{lakshminarayanan2016simple} with its variants (Snapshot Ensemble \citep{snapshotpaper}, Fast Geometric Ensembling (FGE) \citep{geoens}, SWA-Gaussian (SWAG) \citep{maddox2019simple}). 
%~~~~~revised on 2-20~~~~~%
Methods from these two categories thrive in recent years, and a natural idea is to combine them together. Recently, \citet{Ashukha2020Pitfalls} points out the necessity of TS when DE is miscalibrated out-of-the-box. However, due to the intrinsic uncertainty and stochasticity of the learning process, it is possible that the ensemble members are not created equal, and simply averaging over all members may lead to a suboptimal ensemble result. Furthermore, when TS is appended after DE, the original ensemble members are all neglected. As a matter of fact, we still lack a method that can bridge the divide and make the most of the ensemble members.
% **Note: pitfall paper used TS after DE.** Methods from these two categories thrive in recent years, but they are, unfortunately, usually performed separately. Thus, a couple of natural questions arise: Can they be used together to complement each other?  How to make the most of the ensemble members? These questions are yet to be answered.

A plethora of metrics have been developed during the past few years for the evaluation of calibration performance, such as Expected Calibration Error (ECE) \citep{Naeini} and kernel density estimation-based ECE (ECE\textsuperscript{\emph{KDE}}) \citep{mixnmatch}, Log-Likelihood (LL) \citep{guo2017calibration} and calibrated LL \citep{Ashukha2020Pitfalls}, Kolmogorov-Smirnov (KS) test \citep{gupta2021calibration}, etc. Every such metrics has its strengths and weaknesses. For example, ECE could be easily biased by the binning scheme; LLs have also shown themselves to be closely correlated with accuracy ($\rho \thicksim 0.9$) \citep{Ashukha2020Pitfalls}. Moreover, current post-hoc calibration methods usually restrict themselves to only one specific metric (e.g. LL), and it is believed that some metrics (e.g. ECE) can hardly be optimized directly \citep{Ashukha2020Pitfalls}. Thus, there is a crucial need for an optimization framework that allows multiple metrics, including ECEs, to be considered at the same time. 
%into consideration, .
% perhaps the most popular ones. | Since there is no ground-truth regarding the actual uncertainty of the prediction made by a learning system, it is often difficult to empirically measure the uncertainty. 

% ========== our goal and contributions ==========
To address these challenges, we propose in this paper a truth discovery-based framework and an accompanying geometric optimization strategy that is (a) more expressive, (b) metric-agnostic, and (c) beneficial to both ensemble-based and post-hoc calibration of DNNs.

Truth discovery, concerning about finding the most trustworthy information from a number of unreliable sources, is a well-established technique in data mining and theoretic computer science, with firm theoretic foundation \citep{ding2020learning,huang2019faster,li2020approximating}. It finds applications in resolving disagreements from possibly conflicting information sources e.g., crowdsourcing aggregation. In this paper, we intend to answer the following question: \emph{Can truth discovery be used to aggregate information from Deep Ensemble, and in turn to help uncertainty calibration?} This is conceivable because the perturbation within the opinions made by multiple classifiers may reflect the intrinsic uncertainty level of data: if an unanimity of opinion is reached by all classifiers, then the uncertainty level should be relatively low. More importantly, this unanimity has nothing to do with whether the opinions, i.e. predictions, are correct or not. Since the collections of classifiers may provide orthogonal information beyond a single classifier itself, we expect that this information could be unearthed via truth discovery.

Accordingly, in this paper, we propose truth discovery as an ideal tool for improving uncertainty calibration of deep learning, and make several contributions as follows:
\begin{enumerate}
  \item We propose \underline{T}ruth \underline{D}iscovery \underline{E}nsemble (TDE) that improves Deep Ensemble, and show that model uncertainty can be easily derived from truth discovery.
  \item Considering that uncertainty calibration approaches may potentially cause a diminished accuracy, we further develop a provably \underline{a}ccuracy-preserving \underline{T}ruth \underline{D}iscovery \underline{E}nsemble (aTDE) via geometric optimization.
  \item We propose an optimization approach that directly minimizes ECEs, works for both histogram-based and KDE-based metrics, and integrates multiple metrics via compositional training.
  \item We further incorporate the discovered information (i.e. Entropy based Geometric Variance) into the post-hoc calibration pipeline (pTDE) and elevate the performance to a higher level.
\end{enumerate}
%~~~~~redundant?~~~~~
To summarize, we show how truth discovery can benefit both ensemble-based and post-hoc uncertainty calibrations, and validate our proposed methods via experiments upon large-scale datasets, using both binning-based and binning-free metrics, along with comprehensive ablation studies.

% ==================== ==================== ==================== ==================== ====================
% 2 Preliminaries
% ==================== ==================== ==================== ==================== ====================
\section{Preliminaries of Uncertainty Calibration}

For an arbitrary multi-class classifier (not necessarily neural network) $f_{\theta} : \mathcal{D} \subseteq \mathbb{R}^{d} \rightarrow \mathcal{Z} \subseteq \triangle^{L}$ that can make $L$ predictions for $L$ classes, 
%the network 
its outputs (in any scale) can be transformed into a "probability vector" $\mathbf{z} \in \mathcal{Z}$ such that:
\begin{equation}\label{eq:sum}
  \sum_{l=1}^{L} z_l = 1, 0 \leq z_l \leq 1.
\end{equation}
This can be done by the softmax function, which usually tails the last layer of a deep neural network. Here, $\triangle^{L}$ is the probability simplex in $L$ dimensional space. Note that the classifier parameters can also be drawn from a distribution $\theta \thicksim q(\theta)$, e.g., ResNets with random initialization being the only difference.

Although $\mathbf{z}$ is in the probability simplex $\triangle^{L}$, its components may not necessarily have anything to do with, but sometimes are misinterpreted as, the probability of each class. Similarly, the maximum value of the $L$ outputs, $\max_l z_l$, was used to represent the "confidence" that the classifier has on its prediction. 
%In order 
To avoid possible misleading, $\max_l z_l$ is referred to as \emph{winning score} $v$ (i.e., $v = \max_l z_l$) \citep{thulasidasan2019mixup} hereinafter. 

For both ensemble-based and post-hoc calibration methods, the model is trained based on a set of $N_t$ training samples $\{(\mathbf{x}^{(i)}, y^{(i)})\}_{i=1}^{N_t}, \mathbf{x}^{(i)} \in \mathcal{D}, y^{(i)} \in \{1,...,L\}$. Let random variables $X, Y$ represent input data and label, respectively. Then, another random variable $Z = f_{\theta}(X)$ stands for the probability vector. If $z_l$ indeed represents the actual probability of class $l$ (which usually not), then, the following should hold:
\begin{equation}\label{eq:p}
  P(Y = l | Z = \mathbf{z}) = z_l.
\end{equation}
At this time, we also call the classifier $f_{\theta}$ to be perfectly calibrated. It is well known that the probabilities $P(Y = l | Z = \mathbf{z})$ are hard to evaluate, since there is no ground-truth for the probability of an input $\mathbf{x}^{(i)}$ being misclassified as class $l \neq y^{(i)}$. In this paper, we focus on a variant of Eq. (\ref{eq:p}), which only measures the probability of the sample being correctly classified:
\begin{equation}\label{eq:p2}
  P(Y = y^{(i)} | Z = \mathbf{z}^{(i)}) = v^{(i)}
\end{equation}
where $v^{(i)}$ is the winning score which is also the only value taken into consideration when evaluate top-1 accuracy (ACC).

% ~~~~~~~~~~~~~~~~~~~~ ~~~~~~~~~~~~~~~~~~~~ ~~~~~~~~~~~~~~~~~~~~
\paragraph{Ensemble of Deep Neural Networks.} Although it is 
%being 
computationally demanding, ensemble (Deep Ensemble \citep{lakshminarayanan2016simple}, Snapshot Ensemble \citep{snapshotpaper}, etc.) remains as a popular approach for uncertainty calibration.
Formally, for a classifier $f_{\theta}$ with parameter distribution $q(\theta)$, the prediction of sample $\mathbf{x}^{(i)}$ is given by:
\begin{equation}\label{eq:ens}
  \mathbf{z}_{ens}^{(i)} = \int f_{\theta}( \mathbf{x}^{(i)} ) q(\theta) d\theta
\end{equation}
which can be approximated by $S$ independently trained classifiers as $\frac{1}{S} \sum_{s=1}^{S} f_{\theta^{(s)}} (\mathbf{x^{(i)}}), \theta^{(s)} \thicksim q(\theta)$. The $S$ classifiers can be obtained either by independent random initialization (Deep Ensemble) or periodically convergence into local minimum via learning rate decay (Snapshot Ensemble). Since each $\mathbf{z}_{\theta}^{(i)} \in \triangle^{L}$, $\mathbf{z}_{ens}^{(i)}$ is also on the probability simplex. 

% ~~~~~~~~~~~~~~~~~~~~ ~~~~~~~~~~~~~~~~~~~~ ~~~~~~~~~~~~~~~~~~~~
\paragraph{Post-hoc Calibration of Deep Neural Networks.}

Until now, we have not yet introduced the concept of confidence, since all non-post-hoc approaches take the winning score as a representation of the confidence, based on which the calibration error is subsequently measured. In post-hoc calibration, on the other hand, a set of hold-out calibration samples $(\mathbf{x}^{(i)}, y^{(i)})_{i=1}^{N_c}$ is required to learn a mapping from $\mathbf{z}$ to another probability vector $\mathbf{\pi} = \mathcal{T}(\mathbf{z}) $ with a learnable function $ \mathcal{T}: \triangle^L \to \triangle^L$, and $\max_l \pi_l$ is referred to as \emph{confidence} $w$ in this paper, i.e. $w=\max_l \pi_l$. Note that $\operatorname*{arg\,max}_l \pi_l$ is not necessarily equal to $\operatorname*{arg\,max}_l z_l$, and at this time the calibration may potentially decrease the accuracy, if the latter is the correct class. With this, our goal Eq. (\ref{eq:p2}) now becomes $P(Y = y^{(i)} | Z = \mathbf{z}^{(i)}) = w^{(i)}$.
%\begin{equation}\label{eq:p3}
%  P(Y = y^{(i)} | Z = \mathbf{z}^{(i)}) = w^{(i)}.
%\end{equation}
Since we tackle both post-hoc and non-post-hoc calibrations in this paper, we distinguish \emph{winning score} and \emph{confidence} explicitly in that the former directly comes from the classifier $f_{\theta}$ while the latter is derived from a mapping deliberately learned for confidence modeling.

% ~~~~~~~~~~~~~~~~~~~~ ~~~~~~~~~~~~~~~~~~~~ ~~~~~~~~~~~~~~~~~~~~
\paragraph{Calibration Error Evaluation.}

%~~~~~note: using ECE(f_{\theta}) is unclear
For the evaluation of a calibration algorithm, the calibration function $\mathcal{T}$ is applied on another evaluation dataset $(\mathbf{x}^{(i)}, y^{(i)})_{i=1}^{N_e}$ of size $N_e$ which has no overlapping with neither  the training nor the calibration datasets. In histogram-based evaluation metric,  the $N_e$ samples are split into $B$ predefined bins. Formally, we define $B$ pairs of endpoints $\{(\mu_b, \nu_b)\}_{b=1}^B, \nu_b = \mu_{b+1}$, and $B$ point sets $\{P_b\}_{b=1}^{B}: P_b \subseteq \{w_i\}_{i=1}^{N_e}$ such that $\mu_b \leq w < \nu_b, \forall w \in P_b$. Then, the Expected Calibration Error (ECE) is defined as:
\begin{equation}\label{eq:ece}
  ECE(f_{\theta}) = \sum_{b=1}^{B} \frac{|P_b|}{K} |ACC(P_b) - conf(P_b)|,
\end{equation}
which measures the empirical deviation of the sample accuracy in the $b^{th}$ bin: $ACC(P_b) = \frac{1}{|P_b|} \sum_{j=1}^{|P_b|} \bm{1} (\operatorname*{arg\,max}_l z_l^{(j)} = y^{(j)})$ and the average confidence in it: $conf(P_b) = \frac{1}{|P_b|} \sum_{j=1}^{|P_b|} w^{(j)}$. The indicator function $\bm{1} : \mathcal{B} \to \{0, 1\}$ returns 1 if the Boolean expression is true and otherwise 0.

The ECE metric can be easily affected by the number of bins $B$ and the positions of the endpoints. Without the use of binning, ECE\textsuperscript{\emph{KDE}} estimates the calibration error by a kernel function $K : \mathbb{R} \to \mathbb{R}_{\geq 0}$ with bandwidth $h > 0$. Based on Bayesian rule, ECE\textsuperscript{\emph{KDE}} is given as:
\begin{equation}\label{eq:kdeece}
  ECE^{KDE}(f_{\theta}) = \int |w - \widetilde{ACC}(w)| \tilde{P}(w)dw
\end{equation}
in which $\widetilde{ACC}(w)$ is the expected accuracy if the sample confidence is $w$. $\widetilde{P}(w)$ and $\widetilde{ACC}(w)$ are determined by kernel density estimation as: 
\begin{align*} 
  \tilde{P}(w) & = \frac{h^{-1}}{N_e} \sum_{i=1}^{N_e} K_h(w - w^{(i)}), \\ 
  \widetilde{ACC}(w) & = \frac{\sum_{i=1}^{N_e} \bm{1}(\operatorname*{arg\,max}_l z_l^{(i)} = y^{(i)}) K_h(w - w^{(i)})}{\sum_{i=1}^{N_e} K_h(w - w^{(i)})}.
\end{align*}

% ==================== ==================== ==================== ==================== ====================
% 3 Methods I - truth discovery
% ==================== ==================== ==================== ==================== ====================
\section{Truth Discovery Ensebmle}\label{sec:truth}

Existing ensemble techniques in deep learning take the average of predictions made by multiple classifiers 
%that 
derived either from random initialization (Deep Ensemble \citep{lakshminarayanan2016simple}), periodically learning rate decay (Snapshot Ensemble \citep{snapshotpaper}), or from connected optima on the loss functions (Fast Geometric Ensembling \citep{geoens}), but scarcely utilize the sample level variance among the members of an ensemble. 
%has scarcely been utilized. 
To make use of such information, 
%Here 
we first introduce a vanilla truth discovery algorithm in Deep Ensemble context, and then extend it to one with accuracy-preserving guarantee.
%with arguably accuracy-preserving.

% ~~~~~~~~~~~~~~~~~~~~ ~~~~~~~~~~~~~~~~~~~~ ~~~~~~~~~~~~~~~~~~~~
\subsection{Truth Discovery within Probability Simplex}\label{sec:truth-ens} 

To be consistent with truth discovery literature, we use \emph{sources} to denote the $S$ independently trained models. For every sample $(\mathbf{x}^{(i)}, y^{(i)})$ in the evaluation dataset, $S$ independent predictions $\mathbf{z}^{(i, s)} = f_{\theta^{(s)}} (\mathbf{x}^{(i)})$ are made from all $S$ sources (denoted by $\mathbf{z}_s$ hereinafter for brevity). Since the classifiers were trained with stochastic gradient descent (SGD), 
they 
%a classifier 
may 
%possibly 
make wrong decisions 
%upon 
on every $\mathbf{x}^{(i)}$. To model such a behavior,  %so 
we assign a \emph{reliability} value $\omega_s$ to each classifier $f_{\theta^{(s)}}$.

\begin{definition}[Truth discovery \citep{truth14}]
  Given the set of points $\{\mathbf{z}_s\}_{s=1}^{S} \subseteq \triangle^L$ from $S$ classifiers, truth discovery aims at finding the truth probability vector $\mathbf{z}^* \in \triangle^L$ and meanwhile the reliability $\omega_s$ for the $s^{th}$ classifier, such that the following objective function is minimized:  
  \begin{equation}\label{eq:true-min}
    \begin{problem}
      \minimize_{\mathbf{z}^*,\{\omega_s\}_{s=1}^{S}} \quad& \sum_{s=1}^{S} \omega_s ||\mathbf{z}^* - \mathbf{z}_s||^2 \\
      \textrm{s.t.} \quad& \sum_{s=1}^{S} e^{-\omega_s} = 1.
    \end{problem}
  \end{equation}  
\end{definition}
With this definition, interestingly, we can show a direct relationship between truth discovery and model uncertainty. We additionally define \emph{uncertainty of source}, $\upsilon_s$, as the opposite of source reliability, i.e. $\upsilon_s = e^{-\omega_s}$. Then, (\ref{eq:true-min}) can be written as:
\begin{equation}\label{eq:true-min2}
  \begin{problem}
    \minimize_{\mathbf{z}^*,\{\upsilon_s\}_{s=1}^{S}} \quad& \sum_{s=1}^{S} -\frac{||\mathbf{z}^* - \mathbf{z}_s||^2}{2}\ln{\upsilon_s}  \\
    \textrm{s.t.} \quad& \sum_{s=1}^{S} \upsilon_s = 1.
  \end{problem}
\end{equation}  
This is essentially the \emph{Cross Entropy} (CE) of source uncertainty $\upsilon_s$ and $||\mathbf{z}^* - \mathbf{z}_s||/2$, which is the similarity between the optimum probability vector to each source vector ($0 \leq \upsilon_s \leq 1, 0 \leq ||\mathbf{z}^* - \mathbf{z}_s||/2 \leq 1$). Thus, the minimization process is to ensure that the solution resolves the ambiguity of the system as much as possible. Hence, truth discovery can ideally benefit uncertainty calibration through finding the truth vector. 

Algorithms for approximating the global optimum exist \citep{ding2020learning,huang2019faster}.
%li2020approximating}, 
But here with the assumption Eq. (\ref{eq:sum}) that all the possible truth vectors fall on the probability simplex $\triangle^L$, we adopt a simpler solution. Since both the truth vector $\mathbf{z}^*$ and source reliabilities are unknown, we can alternatively 
%iteratively 
update the reliability/uncertainty and the truth vector. Specifically, if $\mathbf{z}^*$ is temporarily fixed, the optimum reliability values can be found through Lemma \ref{lem:fixed}:

\begin{lemma}[\citep{truth14}]\label{lem:fixed}
  If $\mathbf{z}^*$ is fixed, the following reliability value for each source $\omega_s$ minimizes the objective function (\ref{eq:true-min}),
  \begin{equation}\label{eq:updatew}
    \omega_s = \ln(\frac{\sum_{t=1}^{S} ||\mathbf{z}^* - \mathbf{z}_t||^2}{||\mathbf{z}^* - \mathbf{z}_s||}).
  \end{equation}  
\end{lemma}

After the reliabilities have been fixed, the new truth vector can be updated by simply taking the average of source vectors weighted by the found reliabilities, i.e., $\frac{1}{\sum_{s=1}^{S} \omega_s} \sum_{s=1}^{S} \omega_s \mathbf{z}_s$. It can be easily justified that the updated vector is still on the probability simplex. Initially, the ensemble vector $\mathbf{z}_{ens}$ can be an educated guess of $\mathbf{z}^*$. The iterative updating of the truth vector and the source reliability can be terminated if the position of the truth vector changed by less than $\epsilon$ within maximum $I$ iterations. The process is summarized in Algorithm \ref{alg:truth} (ignore line \#5 at this time).

% ~~~~~~~~~~~~~~~~~~~~ ~~~~~~~~~~~~~~~~~~~~ ~~~~~~~~~~~~~~~~~~~~ algorithm: truth finder
\IncMargin{1em}
\begin{algorithm}
  \SetAlgoNoLine
  \KwData{$\{\mathbf{z}_s\}_{s=1}^{S}$}
  \KwResult{$\mathbf{z}^*$}
  $\mathbf{z}^{*(0)} \gets  \mathbf{z}_{ens}$\;
  \For(){$i \gets 1,...,I$}{
    $\omega_s^{(i)} \gets \ln(\sum_{t=1}^{S} ||\mathbf{z}^{*(i-1)} - \mathbf{z}_t||^2 / ||\mathbf{z}^{*(i-1)} - \mathbf{z}_s||)$\;
    $\mathbf{z}^{*(i)} \gets \frac{1}{\sum_{s=1}^{S} \omega_s^{(i)}} \sum_{s=1}^{S} \omega_s^{(i)} \mathbf{z}_s$ \Comment*[r]{update truth vector}
    $\mathbf{z}^{*(i)} \gets \textrm{\textbf{Algorithm} \ref{alg:acc}}(\mathbf{z}^{*(i)})$ \Comment*[r]{preserve accuracy}
    \If{$||\mathbf{z}^{*(i)} - \mathbf{z}^{*(i-1)}||^2 < \epsilon$}{
      \Return{$\mathbf{z}^{*(i)}$}
    }
  }      
  \caption{Optimization of the truth vector.}\label{alg:truth}
  \end{algorithm}
\DecMargin{1em}

\begin{figure}[t]
  \centering
  \includegraphics[width=0.65\linewidth]{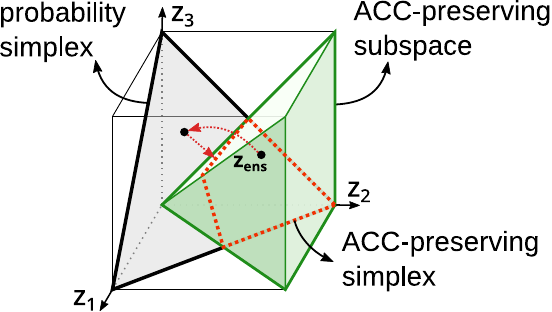}
  \caption{Geometric optimization of accuracy-preserving truth vector in $\mathbb{R}^3$. Here, $2$ is the predicted class, and the accuracy-preserving simplex of class $2$ is highlighted with red boundary.}\label{fig:illus}
\end{figure}

% ~~~~~~~~~~~~~~~~~~~~ ~~~~~~~~~~~~~~~~~~~~ ~~~~~~~~~~~~~~~~~~~~
\subsection{Accuracy-preserving Truth Discovery}\label{sec:truth-acc} 

Post-hoc calibration methods may potentially cause a decrease of the prediction accuracy, if the ranks of the scores for each class cannot be maintained. Hence, we usually anticipate the calibration algorithm to be \emph{accuracy-preserving} \citep{mixnmatch}. In the vanilla truth discovery algorithm, however, this requirement may not be satisfied, since $\mathbf{z}^*$ could change the predictions. Suppose that $c$ is the predicted class derived from ensemble $\mathbf{z}_{ens}$, i.e., $ c = \operatorname*{arg\,max}_l (z_{ens})_l$. After the truth vector $\mathbf{z}^*$ is found, we want to at least maintain the accuracy of ensemble. Thus, Eq. (\ref{eq:true-min}) can be retrofitted to be an accuracy-preserving version.

\begin{definition}[Accuracy-preserving truth discovery]
  Given the set of points $\{\mathbf{z}_s\}_{s=1}^{S} \subseteq \triangle^L$ and the ensemble vector $\mathbf{z}_{ens}$, find the truth vector $\mathbf{z}^* \in \triangle^L$ and reliabilities $\{\omega_s\}_{s=1}^{S}$ such that the following objective function is minimized:  
  \begin{equation}\label{eq:true-min-acc}
    \begin{problem}
      \minimize_{\mathbf{z}^*,\{\omega_s\}_{s=1}^{S}} \quad& \sum_{s=1}^{S} \omega_s ||\mathbf{z}^* - \mathbf{z}_s||^2 \\
      \textrm{s.t.} \quad& \sum_{s=1}^{S} e^{-\omega_s} = 1 \\
      \quad& \operatorname*{arg\,max}_l z_l^* = \operatorname*{arg\,max}_l (z_{ens})_l.
    \end{problem}
  \end{equation}  
\end{definition}
It can be formulated as a geometric optimization problem. See Figure~\ref{fig:illus} for an illustration when $L=3$. The constraint $\operatorname*{arg\,max}_l z_l^* = \operatorname*{arg\,max}_l (z_{ens})_l$ introduces a subspace $\Omega: z_l > z_m, \forall m \neq l$. 
%then 
The discovery of the truth vector must be performed within the accuracy-preserving simplex $\triangle_a = \triangle_L \cap \Omega$. Intuitively, when $\mathbf{z}^*$ falls outside of the accuracy-preserving simplex $\triangle_a$ in one iteration, we can find the projection of $\mathbf{z}^*$ onto $\triangle_a$ to pull it back to $\triangle_a$. This projection can be found by Algorithm \ref{alg:acc} which is proved in Theorem \ref{thm:thm}. When the projection is done, Algorithm \ref{alg:truth} continues until the desired truth vector is found.

\begin{theorem}\label{thm:thm}
  Algorithm \ref{alg:acc} preserves the accuracy of the prediction.
\end{theorem}

\begin{proof}
  The theorem can be proved using Lagrange Multipliers with Karush-Kuhn-Tucker (KKT) Conditions, see supplementary material Sec. \ref{supp:proof} for details.
\end{proof}

% ~~~~~~~~~~~~~~~~~~~~ ~~~~~~~~~~~~~~~~~~~~ ~~~~~~~~~~~~~~~~~~~~ algorithm: acc-preserving
\IncMargin{1em}
\begin{algorithm}
  \SetAlgoNoLine
  \KwData{$\mathbf{z}$}
  \KwResult{$\mathbf{\tilde{z}}$}
  $c \gets \operatorname*{arg\,max}_l (z_{ens})_l;\ \mathbf{\tilde{z}} \gets \mathbf{z}$\;
  $M \gets \var{ARGSORT}(\{z_1,...,z_L\})$\;
  \If{$M[1] = c$}{
    \Return{$\mathbf{z}$}
  }
  \Else(){
    \For(){$l \gets 1,...,L$}{
      $\tilde{z} \gets \frac{1}{l + 1}(z_c + z_{M[1]} + ... + z_{M[l]})$\;
      \If(){$\tilde{z} > z_{M[l+1]}$}{
        $\tilde{z}_c \gets \tilde{z}; \tilde{z}_{M[n]} \gets \tilde{z}, \forall n \leq l$\;
        \Return{$\mathbf{\tilde{z}}$}
      }
    }
  }    
  \caption{Projection from truth vector $\mathbf{z}$ onto accuracy-preserving simplex $\triangle_a$.}\label{alg:acc}
  \end{algorithm}
\DecMargin{1em}

% ~~~~~~~~~~~~~~~~~~~~ ~~~~~~~~~~~~~~~~~~~~ ~~~~~~~~~~~~~~~~~~~~ table: truth begin
% \usepackage{booktabs}
% \usepackage{multirow}
\begin{table*}[!htp]
  \centering  
  \caption{Comparison of TDE/aTDE with Deep Ensemble (DE) before post-hoc calibration.}\label{tab:truth}
  \small
  \resizebox{\textwidth}{!}{%
  \begin{tabular}{@{}lrrrr|rrr|rrrrrr@{\hskip 0.1in}l@{}}
  \toprule
                                 &                      & \multicolumn{6}{c|}{CIFAR100}                                                                                                                                                                                                                   & \multicolumn{6}{c}{CIFAR10}                                                                                                                                                                                                                                          &  \\ \cmidrule(l){3-15} 
                                 & \multicolumn{1}{l}{} & \multicolumn{3}{c|}{50 sources}                                                                                        & \multicolumn{3}{c|}{100 sources}                                                                                       & \multicolumn{3}{c|}{50 sources}                                                                                                             & \multicolumn{3}{c}{100 sources}                                                                                        &  \\ \midrule
  Model                          & Method               & ECE\textsuperscript{\emph{KDE}}$\downarrow$ & ECE$\downarrow$                 & ACC$\uparrow$                          & ECE\textsuperscript{\emph{KDE}}$\downarrow$ & ECE$\downarrow$                 & ACC$\uparrow$                          & ECE\textsuperscript{\emph{KDE}}$\downarrow$ & ECE$\downarrow$                 & \multicolumn{1}{r|}{ACC$\uparrow$}                          & ECE\textsuperscript{\emph{KDE}}$\downarrow$ & ECE$\downarrow$                 & ACC$\uparrow$                          &  \\ \midrule
                                 &DE        & 2.88                                  & 2.50                                  & 82.83                                  & 3.06                                  & 2.58                                  & \textbf{83.02}                         & 1.31                                  & \textbf{0.48}                         & \multicolumn{1}{r|}{\textbf{96.38}}                         & 1.15                                  & \textbf{0.43}                         & 96.41                                  &  \\
                                 & TDE     & \cellcolor[HTML]{EFEFEF}1.60          & \cellcolor[HTML]{EFEFEF}1.98          & \cellcolor[HTML]{EFEFEF}\textbf{82.89} & \cellcolor[HTML]{EFEFEF}1.81          & \cellcolor[HTML]{EFEFEF}\textbf{1.94} & \cellcolor[HTML]{EFEFEF}83.01          & \cellcolor[HTML]{EFEFEF}\textbf{1.05} & \cellcolor[HTML]{EFEFEF}0.75          & \multicolumn{1}{r|}{\cellcolor[HTML]{EFEFEF}\textbf{96.38}} & \cellcolor[HTML]{EFEFEF}\textbf{1.01} & \cellcolor[HTML]{EFEFEF}0.59          & \cellcolor[HTML]{EFEFEF}\textbf{96.42} &  \\
  \multirow{-3}{*}{PreResNet110} & aTDE    & \cellcolor[HTML]{EFEFEF}\textbf{1.55} & \cellcolor[HTML]{EFEFEF}\textbf{1.92} & \cellcolor[HTML]{EFEFEF}82.83          & \cellcolor[HTML]{EFEFEF}\textbf{1.78} & \cellcolor[HTML]{EFEFEF}1.95          & \cellcolor[HTML]{EFEFEF}\textbf{83.02} & \cellcolor[HTML]{EFEFEF}1.07          & \cellcolor[HTML]{EFEFEF}0.75          & \multicolumn{1}{r|}{\cellcolor[HTML]{EFEFEF}\textbf{96.38}} & \cellcolor[HTML]{EFEFEF}1.02          & \cellcolor[HTML]{EFEFEF}0.60          & \cellcolor[HTML]{EFEFEF}96.41          &  \\ \midrule
                                 & DE        & 2.39                                  & 2.09                                  & \textbf{83.52}                         & 2.46                                  & 2.00                                  & \textbf{83.55}                         & 1.32                                  & \textbf{0.34}                         & \multicolumn{1}{r|}{\textbf{96.68}}                         & 1.31                                  & \textbf{0.35}                         & 96.67                                  &  \\
                                 & TDE     & \cellcolor[HTML]{EFEFEF}\textbf{1.31} & \cellcolor[HTML]{EFEFEF}\textbf{1.72} & \cellcolor[HTML]{EFEFEF}83.49          & \cellcolor[HTML]{EFEFEF}1.44          & \cellcolor[HTML]{EFEFEF}\textbf{1.61} & \cellcolor[HTML]{EFEFEF}83.54          & \cellcolor[HTML]{EFEFEF}1.14          & \cellcolor[HTML]{EFEFEF}0.65          & \multicolumn{1}{r|}{\cellcolor[HTML]{EFEFEF}96.66}          & \cellcolor[HTML]{EFEFEF}\textbf{1.08} & \cellcolor[HTML]{EFEFEF}0.52          & \cellcolor[HTML]{EFEFEF}\textbf{96.68} &  \\
  \multirow{-3}{*}{PreResNet164} & aTDE   & \cellcolor[HTML]{EFEFEF}1.33          & \cellcolor[HTML]{EFEFEF}1.76          & \cellcolor[HTML]{EFEFEF}\textbf{83.52} & \cellcolor[HTML]{EFEFEF}\textbf{1.42} & \cellcolor[HTML]{EFEFEF}1.62          & \cellcolor[HTML]{EFEFEF}\textbf{83.55} & \cellcolor[HTML]{EFEFEF}\textbf{1.13} & \cellcolor[HTML]{EFEFEF}0.63          & \multicolumn{1}{r|}{\cellcolor[HTML]{EFEFEF}\textbf{96.68}} & \cellcolor[HTML]{EFEFEF}\textbf{1.08} & \cellcolor[HTML]{EFEFEF}0.53          & \cellcolor[HTML]{EFEFEF}96.67          &  \\ \midrule
                                 & DE        & 6.45                                  & 5.85                                  & \textbf{84.38}                         & 6.40                                  & 5.78                                  & 84.28                                  & \textbf{1.06}                         & \textbf{0.35}                         & \multicolumn{1}{r|}{\textbf{97.17}}                         & 1.20                                  & 0.41                                  & \textbf{97.20}                         &  \\
                                 & TDE     & \cellcolor[HTML]{EFEFEF}\textbf{5.48} & \cellcolor[HTML]{EFEFEF}\textbf{5.03} & \cellcolor[HTML]{EFEFEF}84.37          & \cellcolor[HTML]{EFEFEF}5.58          & \cellcolor[HTML]{EFEFEF}5.07          & \cellcolor[HTML]{EFEFEF}\textbf{84.29} & \cellcolor[HTML]{EFEFEF}1.14          & \cellcolor[HTML]{EFEFEF}0.49          & \multicolumn{1}{r|}{\cellcolor[HTML]{EFEFEF}97.15}          & \cellcolor[HTML]{EFEFEF}\textbf{1.10} & \cellcolor[HTML]{EFEFEF}\textbf{0.40} & \cellcolor[HTML]{EFEFEF}97.17          &  \\
  \multirow{-3}{*}{WideResNet}   & aTDE    & \cellcolor[HTML]{EFEFEF}5.49          & \cellcolor[HTML]{EFEFEF}5.05          & \cellcolor[HTML]{EFEFEF}\textbf{84.38} & \cellcolor[HTML]{EFEFEF}\textbf{5.57} & \cellcolor[HTML]{EFEFEF}\textbf{5.06} & \cellcolor[HTML]{EFEFEF}84.28          & \cellcolor[HTML]{EFEFEF}1.13          & \cellcolor[HTML]{EFEFEF}0.47          & \multicolumn{1}{r|}{\cellcolor[HTML]{EFEFEF}\textbf{97.17}} & \cellcolor[HTML]{EFEFEF}1.12          & \cellcolor[HTML]{EFEFEF}0.41          & \cellcolor[HTML]{EFEFEF}\textbf{97.20} &  \\ \midrule
                                 &                      & \multicolumn{6}{c|}{ImageNet}                                                                                                                                                                                                                   & \multicolumn{6}{c}{ImageNet (Snapshot Ensemble)}                                                                                                                                                                                                                     &  \\ \cmidrule(l){3-15} 
                                 & \multicolumn{1}{l}{} & \multicolumn{3}{c|}{25 sources}                                                                                        & \multicolumn{3}{c|}{50 sources}                                                                                        & \multicolumn{3}{c|}{25 sources}                                                                                                             & \multicolumn{3}{c}{50 sources}                                                                                         &  \\ \midrule
                                 & DE/SE        & 3.11                                  & 3.12                                  & \textbf{79.25}                         & 3.24                                  & 3.17                                  & \textbf{79.37}                         & 1.85                                  & 2.11                                  & \multicolumn{1}{r|}{\textbf{78.46}}                         & 1.73                                  & \textbf{2.03}                         & \textbf{78.52}                         &  \\
                                 & TDE     & \cellcolor[HTML]{EFEFEF}\textbf{2.16} & \cellcolor[HTML]{EFEFEF}\textbf{2.42} & \cellcolor[HTML]{EFEFEF}79.22          & \cellcolor[HTML]{EFEFEF}\textbf{2.41} & \cellcolor[HTML]{EFEFEF}\textbf{2.58} & \cellcolor[HTML]{EFEFEF}79.35          & \cellcolor[HTML]{EFEFEF}\textbf{1.69} & \cellcolor[HTML]{EFEFEF}\textbf{2.08} & \multicolumn{1}{r|}{\cellcolor[HTML]{EFEFEF}78.45}          & \cellcolor[HTML]{EFEFEF}\textbf{1.66} & \cellcolor[HTML]{EFEFEF}2.10          & \cellcolor[HTML]{EFEFEF}78.50          &  \\
  \multirow{-3}{*}{ResNet50}     & aTDE    & \cellcolor[HTML]{EFEFEF}2.19          & \cellcolor[HTML]{EFEFEF}2.45          & \cellcolor[HTML]{EFEFEF}\textbf{79.25} & \cellcolor[HTML]{EFEFEF}2.43          & \cellcolor[HTML]{EFEFEF}2.60          & \cellcolor[HTML]{EFEFEF}\textbf{79.37} & \cellcolor[HTML]{EFEFEF}1.70          & \cellcolor[HTML]{EFEFEF}2.10          & \multicolumn{1}{r|}{\cellcolor[HTML]{EFEFEF}\textbf{78.46}} & \cellcolor[HTML]{EFEFEF}1.69          & \cellcolor[HTML]{EFEFEF}2.13          & \cellcolor[HTML]{EFEFEF}\textbf{78.52} &  \\ \bottomrule
  \end{tabular}%
  }
\end{table*}
% ~~~~~~~~~~~~~~~~~~~~ ~~~~~~~~~~~~~~~~~~~~ ~~~~~~~~~~~~~~~~~~~~ table: truth end

% ==================== ==================== ==================== ==================== ====================
% 4 Methods II - Optimization
% ==================== ==================== ==================== ==================== ====================
\section{Truth Discovery-regularized Post-hoc Calibration}\label{sec:opt} 

Although  ECE-like scores are difficult to be optimized directly, recent works have attempted to minimize ECE either by using maximum mean calibration error (MMCE), a kernelized version of the ECE \citep{KumarTrainable}, or by rank preserving transforms \citep{bai2021improved}. To provide a better solution, 
%Here 
we formulate the minimization of ECE (and also ECE\textsuperscript{\emph{KDE}}) as an optimization problem in high dimensions, and show how it can be easily extended to incorporate the information gained from truth discovery.

% ~~~~~~~~~~~~~~~~~~~~ ~~~~~~~~~~~~~~~~~~~~ ~~~~~~~~~~~~~~~~~~~~
\paragraph{Optimization of ECEs.}\label{para:opt-hist-ece} 

For simplicity, we only consider the confidence for the ground-truth (namely, top-1) class, which is essentially the probability of a sample being correctly predicted. Then, our learnable mapping becomes $w = \mathcal{T}(\mathbf{z}): \triangle^L \to \mathbb{R}$. One step further, if only the \emph{winning score} is considered, then $w = \mathcal{T}(v): \mathbb{R} \to \mathbb{R}$. Next, we find the specific form of $\mathcal{T}$. It has been shown that deep neural networks tends to be overconfident on most of the predictions \citep{guo2017calibration}. Inspired by this, we impose an attenuation factor $\varphi(v)$ on every sample, which is a function of the winning score $v$ so that the adjusted confidence becomes $w = v - \varphi(v)$. The simplest form of $\varphi(v)$ is to use a constant within a bin $P_b$. Thus, we define an attenuation weight $\psi_b$ for the bin $P_b$. All the attenuation weights $\{\psi_b\}_{b=1}^{B}$ can be viewed as a point $\mathbf{\psi} \in \mathbb{R}^B$. Now we have the definition of the mapping function:
\begin{gather}\label{eq:factor}
  w  = v - \varphi(v) = v - \psi_{\kappa}\\
  \textrm{s.t.} \quad \mu_{\kappa} \leq v < \nu_{\kappa}.
\end{gather}
Notice that for consistency we call $\varphi(v)$ the attenuation factor, but it can also enhance the confidence $w$ if $\varphi(v) < 0$ somewhere.

%~~~~~ using $ECE(f_\theta)$ is unclear
Now our goal is to find the location of $\mathbf{\psi}$ in $\mathbb{R}^B$ such that the expected calibration error is minimized:
\begin{equation}\label{eq:optece}
  \begin{problem}
    \minimize_{\{\psi_b\}_{b=1}^{B}} \quad&  ECE(\{P_b\}_{b=1}^{B}), \\
  \end{problem}
\end{equation}  
where $ECE$ is shown in Eq. (\ref{eq:ece}). Since all the computations in ECE (and ECE\textsuperscript{\emph{KDE}}) are differentiable, the minimization of ECEs can be done by gradient descent methods. Here, a mini-batch Stochastic Gradient Descent (SGD) approach is used. In each epoch, a subset of calibration data is sampled, based on which the attenuation weights are updated. The optimization process is encapsulated in Algorithm \ref{alg:optece}, in which ECE and ECE\textsuperscript{\emph{KDE}} can be used interchangeably. The algorithm can be easily implemented by virtue of automatic differentiation libraries e.g. PyTorch \citep{pytorchpaper}.

% ~~~~~~~~~~~~~~~~~~~~ ~~~~~~~~~~~~~~~~~~~~ ~~~~~~~~~~~~~~~~~~~~ algorithm: optimization
\IncMargin{1em}
\begin{algorithm}
  \SetAlgoNoLine
  \KwData{$\{(v^{(i)},y^{(i)})\}_{i=1}^{N_c}$}
  \KwResult{$\{\psi_b\}_{b=1}^{B}$} 
  \For(){$\textrm{epoch} \gets 1,...,\textrm{\#epoch}$}{
    sample $\{(v^{(j)},y^{(j)})\}_{j=1}^{n_c} \thicksim \{(v^{(i)},y^{(i)})\}_{i=1}^{N_c}$\;
    $w = v - \varphi(v)$ \Comment*[r]{apply attenuation factor}
    loss $\gets$ ECE or ECE\textsuperscript{\emph{KDE}} \Comment*[r]{forward propagation}
    update $\{\psi_b\}_{b=1}^{B}$ \Comment*[r]{backward propagation}    
  }      
  \Return{$\{\psi_b\}_{b=1}^{B}$}    
  \caption{Optimization of ECEs.}\label{alg:optece}
  \end{algorithm}
\DecMargin{1em}

% ~~~~~~~~~~~~~~~~~~~~ ~~~~~~~~~~~~~~~~~~~~ ~~~~~~~~~~~~~~~~~~~~
\paragraph{Compositional Approach for the Optimization of ECEs.}\label{para:opt-compos} 

Even an accelerated method for KDE computation is used \citep{OBRIEN2016148}, KDE-based metric is still much more time-consuming compared to histogram-based metrics. A natural question that arises here is whether the minimization of ECE\textsuperscript{\emph{KDE}} can be speeded up by the minimization of ECE. To answer this, we first find the attenuation weights by using ECE as the loss function, and then use the obtained $\{\psi_b\}_{b=1}^{B}$ as an initial guess for the minimization of ECE\textsuperscript{\emph{KDE}}. This approach enables the compositional optimization of histogram-based and KDE-based calibration errors.

% ~~~~~~~~~~~~~~~~~~~~ ~~~~~~~~~~~~~~~~~~~~ ~~~~~~~~~~~~~~~~~~~~
\paragraph{ECE Optimization Regularized by Truth Discovery.}\label{para:opt-truth} 

Given a discovered truth vector $\mathbf{z}^*$, let $V$ denote the total squared distance to $\mathbf{z}^*$ (i.e., $V = \sum_{s=1}^{S} ||\mathbf{z}^* - \mathbf{z}_{s}||^2$) and $q_s$ denote the contribution of each $\mathbf{z}_{s}$ to $V$ (i.e., $q_s = ||\mathbf{z}^* - \mathbf{z}_{s}||^2/V$). Then, the entropy induced by $\{q_s\}_{s=1}^{S}$ is:
\begin{equation}\label{eq:hv}
  H = -\sum_{s=1}^{S} q_s \log q_s = \frac{1}{V} \sum_{s=1}^{S} ||\mathbf{z}^* - \mathbf{z}_{s}||^2 \log \frac{V}{||\mathbf{z}^* - \mathbf{z}_{s}||^2}.
\end{equation} 
Based on these, we can define the \emph{Entropy based Geometric Variance} ($HV$):

\begin{definition}[Entropy based Geometric Variance \citep{ding2020learning}]
  Given the point set $\{\mathbf{z}_s\}_{s=1}^{S}$ and a point $\mathbf{z}^*$, the Entropy based Geometric Variance ($HV$) is $H \times V$ where $H$ and $V$ are defined as shown above. 
\end{definition}

With this definition, it is easy to see that the objective function of truth discovery (\ref{eq:true-min}) is exactly the entropy based geometric variance ($HV$) and the optimization problem (\ref{eq:true-min}) is equivalent to finding a point $\mathbf{z}^*$ to minimize $HV$. 

If the truth vector $\mathbf{z}^*$ has been determined, then $HV$ is an indicator of the ambiguity of the system, and can be borrowed as an external information for our ECE optimization. Despite being in the same bin and overconfident, the sample confidences should not be attenuated at the same scale. Instead, the sample with higher $HV$ (i.e. higher variance and uncertainty) is to be attenuated by a larger magnitude. Consequently, the mapping function can be reshaped as:
\begin{gather}\label{eq:factor-hv}
  w^{(i)} = v^{(i)} - \varphi(v^{(i)}) = v^{(i)} - \alpha_1 \psi_{\kappa} - \alpha_2 HV^{(i)},
\end{gather}
where $\alpha_1$, $\alpha_2$ are hyperparameters, and $HV^{(i)}$ is essentially the value of the objective function of truth discovery as computed in Section \ref{sec:truth-acc} for each sample $(x^{(i)},y^{(i)})$ in a total of $N_c+N_e$ calibration and evaluation samples. The learning of the mapping function $\mathcal{T}$ from the calibration data is a supervised learning problem. Hence, $\mathcal{T}$ is expected to be overfitted to calibration data. By incorporating orthogonal information (i.e. $HV$) acquired from the truth discovery of multiple ensemble classifiers, we can learn a mapping $\mathcal{T}$ that generalizes better on the evaluation datasets.

% ~~~~~~~~~~~~~~~~~~~~ ~~~~~~~~~~~~~~~~~~~~ ~~~~~~~~~~~~~~~~~~~~ figure: truth main
\begin{figure*}[!htp]
  \centering
  \includegraphics[width=\linewidth]{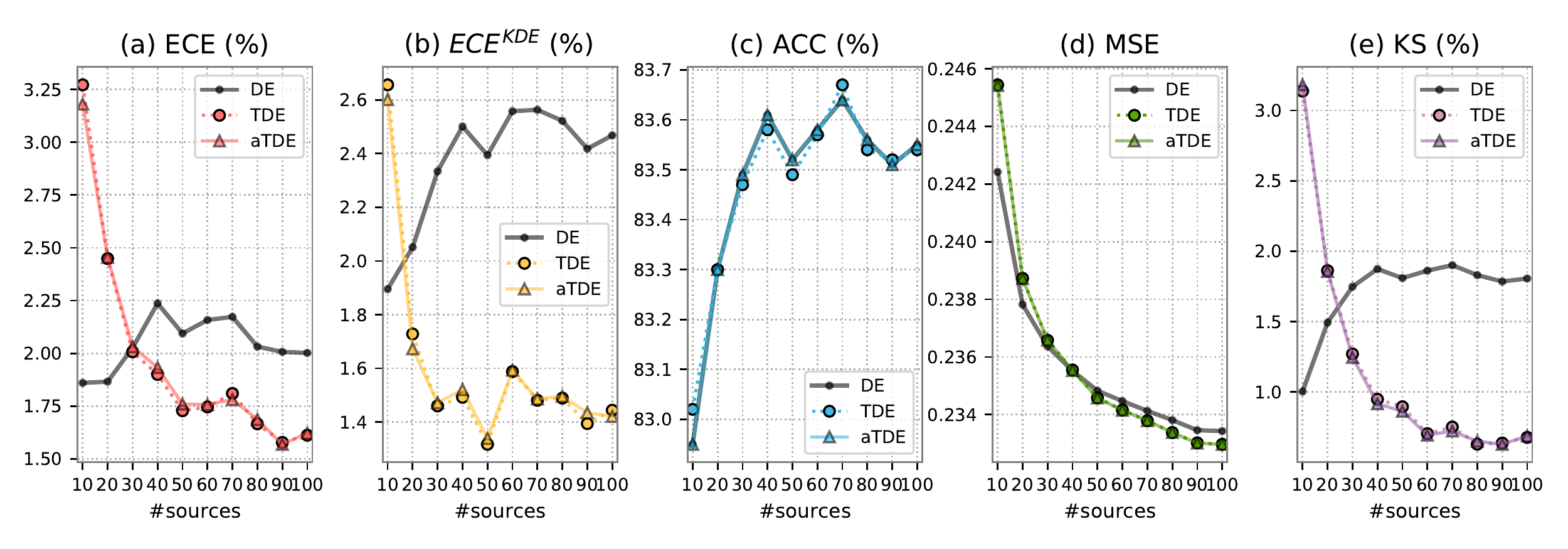}
  \caption{Results of DE/TDE/aTDE on PreResNet164 trained upon CIFAR100 in an unsupervised manner (i.e. no hold-out calibration set). The number of sources is increased from 10 to 100. Our method works favorably with various metrics for the evaluation of calibration.}\label{fig:truth}
\end{figure*}

% ~~~~~~~~~~~~~~~~~~~~ ~~~~~~~~~~~~~~~~~~~~ ~~~~~~~~~~~~~~~~~~~~ table: opt begin
\begin{table*}[!htp]\centering
  \caption{Comparison of post-hoc calibration methods. The \textbf{best} results are highlighted in bold. We also underline the best results \underline{excluding} our method. ECEs are reported in terms of mean$\pm$standard deviation obtained from 5 random replications.}\label{tab:geoC} 
  \resizebox{\textwidth}{!}{%
  \small
  \begin{tabular}{llrrrrr|rrrrrr}\toprule
  & &\multicolumn{5}{c}{ECE$\downarrow$} &\multicolumn{5}{c}{ECE\textsuperscript{\emph{KDE}}$\downarrow$} \\\cmidrule{3-12}
  Dataset &Model &DE &TS &ETS &IRM &pTDE &DE &TS &ETS &IRM &pTDE \\\midrule
  CIFAR100 &ResNet18 &2.86{\scriptsize $\pm$0.42} &2.70{\scriptsize $\pm$0.36} &\underline{2.31{\scriptsize $\pm$0.30}} &2.52{\scriptsize $\pm$0.35} &\cellcolor[HTML]{f3f3f3}\textbf{1.64{\scriptsize $\pm$0.37}} &2.41{\scriptsize $\pm$0.35} &2.37{\scriptsize $\pm$0.37} &2.25{\scriptsize $\pm$0.33} &\underline{2.09{\scriptsize $\pm$0.36}} &\cellcolor[HTML]{f3f3f3}\textbf{1.63{\scriptsize $\pm$0.37}} \\
  CIFAR100 &DenseNet121 &1.69{\scriptsize $\pm$0.15} &1.68{\scriptsize $\pm$0.14} &\underline{1.45{\scriptsize $\pm$0.15}} &1.70{\scriptsize $\pm$0.19} &\cellcolor[HTML]{f3f3f3}\textbf{1.29{\scriptsize $\pm$0.19}} &1.67{\scriptsize $\pm$0.07} &1.66{\scriptsize $\pm$0.08} &\underline{1.65{\scriptsize $\pm$0.12}} &1.66{\scriptsize $\pm$0.15} &\cellcolor[HTML]{f3f3f3}\textbf{1.49{\scriptsize $\pm$0.12}} \\
  CIFAR100 &ResNeXt29 &1.92{\scriptsize $\pm$0.26} &1.93{\scriptsize $\pm$0.27} &\underline{1.54{\scriptsize $\pm$0.22}} &1.88{\scriptsize $\pm$0.36} &\cellcolor[HTML]{f3f3f3}\textbf{1.20{\scriptsize $\pm$0.34}} &1.89{\scriptsize $\pm$0.12} &1.86{\scriptsize $\pm$0.07} &1.87{\scriptsize $\pm$0.11} &\underline{1.74{\scriptsize $\pm$0.24}} &\cellcolor[HTML]{f3f3f3}\textbf{1.63{\scriptsize $\pm$0.12}} \\ \midrule
  CIFAR10 &ResNet18 &0.94{\scriptsize $\pm$0.14} &\underline{0.65{\scriptsize $\pm$0.13}} &\underline{0.65{\scriptsize $\pm$0.13}} &0.67{\scriptsize $\pm$0.24} &\cellcolor[HTML]{f3f3f3}0.75{\scriptsize $\pm$0.10} &1.50{\scriptsize $\pm$0.11} &1.48{\scriptsize $\pm$0.06} &1.48{\scriptsize $\pm$0.06} &\underline{1.35{\scriptsize $\pm$0.08}} &\cellcolor[HTML]{f3f3f3}\textbf{1.21{\scriptsize $\pm$0.18}} \\
  CIFAR10 &DenseNet121 &0.88{\scriptsize $\pm$0.11} &0.65{\scriptsize $\pm$0.13} &0.65{\scriptsize $\pm$0.13} &\underline{0.55{\scriptsize $\pm$0.06}} &\cellcolor[HTML]{f3f3f3}\textbf{0.50{\scriptsize $\pm$0.14}} &1.78{\scriptsize $\pm$0.17} &1.76{\scriptsize $\pm$0.16} &1.76{\scriptsize $\pm$0.16} &\underline{1.56{\scriptsize $\pm$0.15}} &\cellcolor[HTML]{f3f3f3}\textbf{1.45{\scriptsize $\pm$0.20}} \\
  CIFAR10 &ResNeXt29 &0.46{\scriptsize $\pm$0.09} &\underline{\textbf{0.33{\scriptsize $\pm$0.10}}} &0.34{\scriptsize $\pm$0.10} &0.45{\scriptsize $\pm$0.13} &\cellcolor[HTML]{f3f3f3}0.39{\scriptsize $\pm$0.15} &\underline{1.38{\scriptsize $\pm$0.06}} &1.43{\scriptsize $\pm$0.09} &1.43{\scriptsize $\pm$0.09} &1.42{\scriptsize $\pm$0.10} &\cellcolor[HTML]{f3f3f3}\textbf{1.36{\scriptsize $\pm$0.06}} \\ \midrule
  &\#sources &DE/SE &TS &ETS &IRM &pTDE &DE/SE &TS &ETS &IRM &pTDE \\ \midrule
  ImageNet &10 (DE) &3.06{\scriptsize $\pm$0.12} &1.59{\scriptsize $\pm$0.09} &\underline{0.96{\scriptsize $\pm$0.07}} &1.93{\scriptsize $\pm$0.16} &\cellcolor[HTML]{f3f3f3}\textbf{0.88{\scriptsize $\pm$0.11}} &3.13{\scriptsize $\pm$0.11} &1.16{\scriptsize $\pm$0.14} &\underline{1.06{\scriptsize $\pm$0.11}} &1.79{\scriptsize $\pm$0.11} &\cellcolor[HTML]{f3f3f3}\textbf{{0.93\scriptsize $\pm$0.08}} \\
  ImageNet &30 (DE) &3.19{\scriptsize $\pm$0.06} &1.47{\scriptsize $\pm$0.15} &\underline{0.87{\scriptsize $\pm$0.10}} &1.89{\scriptsize $\pm$0.12} &\cellcolor[HTML]{f3f3f3}\textbf{{0.81\scriptsize $\pm$0.16}} &3.26{\scriptsize $\pm$0.05} &1.10{\scriptsize $\pm$0.12} &\underline{0.97{\scriptsize $\pm$0.09}} &1.75{\scriptsize $\pm$0.07} &\cellcolor[HTML]{f3f3f3}\textbf{0.95{\scriptsize $\pm$0.11}} \\
  ImageNet &10 (SE) &2.32{\scriptsize $\pm$0.09} &1.56{\scriptsize $\pm$0.05} &\underline{1.04{\scriptsize $\pm$0.10}} &1.74{\scriptsize $\pm$0.04} &\cellcolor[HTML]{f3f3f3}\textbf{0.71{\scriptsize $\pm$0.17}} &2.18{\scriptsize $\pm$0.09} &1.14{\scriptsize $\pm$0.06} &\underline{1.07{\scriptsize $\pm$0.08}} &1.51{\scriptsize $\pm$0.07} &\cellcolor[HTML]{f3f3f3}\textbf{0.95{\scriptsize $\pm$0.07}} \\
  ImageNet &30 (SE) &2.13{\scriptsize $\pm$0.07} &1.57{\scriptsize $\pm$0.07} &\underline{0.95{\scriptsize $\pm$0.15}} &1.70{\scriptsize $\pm$0.07} &\cellcolor[HTML]{f3f3f3}\textbf{0.76{\scriptsize $\pm$0.22}} &1.89{\scriptsize $\pm$0.09} &1.16{\scriptsize $\pm$0.08} &\underline{0.95{\scriptsize $\pm$0.17}} &1.43{\scriptsize $\pm$0.11} &\cellcolor[HTML]{f3f3f3}\textbf{0.87{\scriptsize $\pm$0.22}} \\
  \bottomrule
  \end{tabular}}
\end{table*}
% ~~~~~~~~~~~~~~~~~~~~ ~~~~~~~~~~~~~~~~~~~~ ~~~~~~~~~~~~~~~~~~~~ table: opt end

% ~~~~~~~~~~~~~~~~~~~~ ~~~~~~~~~~~~~~~~~~~~ ~~~~~~~~~~~~~~~~~~~~ table: ablation begin
\begin{table*}[!htp]\centering
    \caption{Ablation Study of the proposed truth discovery-regularized post-hoc calibration (pTDE). For the four variants of pTDE, the blue/red color denotes if compositional training (Comp.)/truth-discovery regularization (Truth. Reg.) is utilized.}\label{tab:ablation}
    \resizebox{\textwidth}{!}{%
    \small
    \begin{tabular}{llrrrr|rrrrr}\toprule
    & &\multicolumn{4}{c|}{ECE$\downarrow$} &\multicolumn{4}{c}{ECE\textsuperscript{\emph{KDE}}$\downarrow$} \\\cmidrule{3-10}
    Dataset &Model &opt\textsuperscript{\emph{hist}} &opt\textsuperscript{\emph{KDE}} &pTDE\textsuperscript{\emph{hist}} &pTDE\textsuperscript{\emph{KDE}} &opt\textsuperscript{\emph{hist}} &opt\textsuperscript{\emph{KDE}} &pTDE\textsuperscript{\emph{hist}} &pTDE\textsuperscript{\emph{KDE}} \\\midrule
    \multicolumn{2}{c}{\colorbox{blue!30}{\textbf{Comp.}} \colorbox{red!30}{\textbf{Truth. Reg.}}} &\colorbox{blue!30}{\xmark}\colorbox{red!30}{\xmark} &\colorbox{blue!30}{\cmark}\colorbox{red!30}{\xmark} &\colorbox{blue!30}{\xmark}\colorbox{red!30}{\cmark} &\colorbox{blue!30}{\cmark}\colorbox{red!30}{\cmark} &\colorbox{blue!30}{\xmark}\colorbox{red!30}{\xmark} &\colorbox{blue!30}{\cmark}\colorbox{red!30}{\xmark} &\colorbox{blue!30}{\xmark}\colorbox{red!30}{\cmark} &\colorbox{blue!30}{\cmark}\colorbox{red!30}{\cmark} \\
    CIFAR100 &ResNet18 &1.74{\scriptsize $\pm$0.22} &1.68{\scriptsize $\pm$0.38} &\textbf{1.59{\scriptsize $\pm$0.37}} &1.64{\scriptsize $\pm$0.37} &1.76{\scriptsize $\pm$0.33} &1.73{\scriptsize $\pm$0.31} &1.71{\scriptsize $\pm$0.35} &\textbf{1.63{\scriptsize $\pm$0.37}} \\
    CIFAR100 &DenseNet121 &\textbf{1.29{\scriptsize $\pm$0.17}} &1.31{\scriptsize $\pm$0.15} &1.29{\scriptsize $\pm$0.20} &1.29{\scriptsize $\pm$0.19} &1.56{\scriptsize $\pm$0.09} &1.53{\scriptsize $\pm$0.11} &1.52{\scriptsize $\pm$0.09} &\textbf{1.49{\scriptsize $\pm$0.12}} \\
    CIFAR100 &ResNeXt29 &1.27{\scriptsize $\pm$0.33} &1.24{\scriptsize $\pm$0.35} &\textbf{1.19{\scriptsize $\pm$0.35}} &1.20{\scriptsize $\pm$0.34} &1.68{\scriptsize $\pm$0.12} &1.67{\scriptsize $\pm$0.12} &1.64{\scriptsize $\pm$0.11} &\textbf{1.63{\scriptsize $\pm$0.12}} \\ \midrule
    CIFAR10 &ResNet18 &0.65{\scriptsize $\pm$0.20} &0.72{\scriptsize $\pm$0.15} &\textbf{0.65{\scriptsize $\pm$0.17}} &0.75{\scriptsize $\pm$0.10} &1.33{\scriptsize $\pm$0.12} &1.26{\scriptsize $\pm$0.18} &1.31{\scriptsize $\pm$0.12} &\textbf{1.21{\scriptsize $\pm$0.18}} \\
    CIFAR10 &DenseNet121 &0.45{\scriptsize $\pm$0.06} &0.46{\scriptsize $\pm$0.16} &\textbf{0.44{\scriptsize $\pm$0.04}} &0.50{\scriptsize $\pm$0.14} &1.54{\scriptsize $\pm$0.16} &1.46{\scriptsize $\pm$0.17} &1.55{\scriptsize $\pm$0.18} &\textbf{1.45{\scriptsize $\pm$0.20}} \\
    CIFAR10 &ResNeXt29 &0.41{\scriptsize $\pm$0.10} &0.41{\scriptsize $\pm$0.13} &0.40{\scriptsize $\pm$0.11} &\textbf{0.39{\scriptsize $\pm$0.15}} &1.37{\scriptsize $\pm$0.07} &\textbf{1.36{\scriptsize $\pm$0.07}} &1.37{\scriptsize $\pm$0.06} &1.36{\scriptsize $\pm$0.06} \\ \midrule
    &\#sources &opt\textsuperscript{\emph{hist}} &opt\textsuperscript{\emph{KDE}} &pTDE\textsuperscript{\emph{hist}} &pTDE\textsuperscript{\emph{KDE}} &opt\textsuperscript{\emph{hist}} &opt\textsuperscript{\emph{KDE}} &pTDE\textsuperscript{\emph{hist}} &pTDE\textsuperscript{\emph{KDE}} \\ \midrule
    ImageNet &10 (DE) &\textbf{0.76{\scriptsize $\pm$0.13}} &0.83{\scriptsize $\pm$0.13} &0.81{\scriptsize $\pm$0.09} &0.88{\scriptsize $\pm$0.11} &0.99{\scriptsize $\pm$0.08} &\textbf{0.92{\scriptsize $\pm$0.07}} &1.01{\scriptsize $\pm$0.08} &0.93{\scriptsize $\pm$0.08} \\
    ImageNet &30 (DE) &0.88{\scriptsize $\pm$0.13} &0.85{\scriptsize $\pm$0.11} &0.87{\scriptsize $\pm$0.16} &\textbf{0.81{\scriptsize $\pm$0.16}} &1.09{\scriptsize $\pm$0.12} &1.02{\scriptsize $\pm$0.09} &1.03{\scriptsize $\pm$0.10} &\textbf{0.95{\scriptsize $\pm$0.11}} \\
    ImageNet &10 (SE) &0.80{\scriptsize $\pm$0.19} &0.73{\scriptsize $\pm$0.19} &0.75{\scriptsize $\pm$0.19} &\textbf{0.71{\scriptsize $\pm$0.17}} &1.04{\scriptsize $\pm$0.12} &0.96{\scriptsize $\pm$0.09} &1.01{\scriptsize $\pm$0.11} &\textbf{0.95{\scriptsize $\pm$0.07}} \\
    ImageNet &30 (SE) &0.81{\scriptsize $\pm$0.16} &\textbf{0.76{\scriptsize $\pm$0.20}} &0.78{\scriptsize $\pm$0.17} &0.76{\scriptsize $\pm$0.22} &0.98{\scriptsize $\pm$0.19} &0.90{\scriptsize $\pm$0.22} &0.95{\scriptsize $\pm$0.22} &\textbf{0.87{\scriptsize $\pm$0.22}} \\
    \bottomrule
    \end{tabular}}
  \end{table*}
% ~~~~~~~~~~~~~~~~~~~~ ~~~~~~~~~~~~~~~~~~~~ ~~~~~~~~~~~~~~~~~~~~ table: ablation end

% ==================== ==================== ==================== ==================== ====================
% 5 Experiments
% ==================== ==================== ==================== ==================== ====================
\section{Experiments}\label{sec:exp} 

The main goals of our experiments are to: (1) compare Truth Discovery Ensemble (TDE), especially the accuracy-preserving version (aTDE), with ensemble-based calibration methods; (2) for post-hoc calibration scheme, compare truth discovery-regularized calibration (pTDE) with state-of-the-art methods on a wide range of network architectures and datasets; investigate if different components of our methods collaboratively contribute to the overall elevation of performance by ablation studies.

% ~~~~~~~~~~~~~~~~~~~~ ~~~~~~~~~~~~~~~~~~~~ ~~~~~~~~~~~~~~~~~~~~  Result: Part I
\subsection{Improved Deep Ensemble by Truth Discovery}\label{sec:exp-truth} 

\paragraph{Experimental Setup.}
For a fair comparison, we downloaded the trained models\footnote{downloaded from \url{https://github.com/bayesgroup/pytorch-ensembles}.}\label{foot:download} of \citet{Ashukha2020Pitfalls} including PreResNet110/164 \citep{ResNet} and WideResNet28x10 \citep{WideResNet} trained on CIFAR10/100 \citep{CIFAR} (10/100 classes), and ResNet50 trained on ImageNet \citep{ImageNet} (10000 classes). All 3 network architectures on CIFAR10/100 were trained 100 times (i.e. $S=100$) following the Deep Ensemble (DE) workflow, while ResNet50 was trained on ImageNet resulting in 50 models either by Deep Ensemble or by Snapshot Ensemble (i.e. $S=50$). For every sample in the standard testing dataset, $S$ ensemble members were generated from the $S$ models. While looking for the truth vector, for both the vanilla Truth Discovery Ensemble (TDE) and its accuracy-preserving counterpart (aTDE), we set $\epsilon = e^{-8}$ in all experiments, and observed a convergence within typically 5 iterations. 

\paragraph{Results.}
The comparison of truth discovery ensemble methods with Deep Ensemble on CIFARs ($S=50$ or $100$), and with Deep Ensemble/Snapshot Ensemble on ImageNet ($S=25$ or $50$) is shown in Table \ref{tab:truth}. Clearly, TDE ameliorates either ECE or ECE\textsuperscript{\emph{KDE}} by a large margin in most of the experimental settings, especially on datasets with higher complexity (i.e., CIFAR100 and ImageNet), but fails at maintain accuracy. The accuracy-preserving version aTDE, on the other hand, successfully preserves the accuracy, with nearly the same capability of lowering the ECEs, which validates the correctness of our accuracy-preserving Algorithm \ref{alg:acc}. It can also be concluded from Table \ref{tab:truth} that higher ACC and lower ECEs are hard to be reached simultaneously, but the metrics contributed to by the two variants TDE/aTDE are usually very similar. The KS metric recently proposed by \citep{gupta2021calibration} which measures the maximal distance between the accumulated output probability to the actual probability, is a binning-free calibration evaluator that different from ECEs. The KS error is also measured for all the experiments. By leveraging truth discovery, our TDE method lowers the KS to as low as ~0.6\% (100 sources) as shown in Figure \ref{fig:truth}(e), even without any hold-out calibration sample.

Further, we investigate the stability of TDE/aTDE with different number of sources by changing $S$ by an interval of 10 for CIFARs and 5 for ImageNet, as illustrated in Figure \ref{fig:truth} and Tables \ref{tab:supp-truth1}, \ref{tab:supp-truth2}, \ref{tab:supp-truth3}, and \ref{tab:supp-truth4}. Interestingly, Deep Ensemble tends to be overconfident with larger number of sources, i.e., ensemble members, while TDE/aTDE works favorably with even larger amount of available sources, and this is when a high accuracy is usually reached (Figure \ref{fig:truth}c), suggesting TDE and aTDE's superior ability in utilizing information from multiple sources than Deep Ensemble.

% ~~~~~~~~~~~~~~~~~~~~ ~~~~~~~~~~~~~~~~~~~~ ~~~~~~~~~~~~~~~~~~~~ Result: Part II
\subsection{Improved Post-hoc Calibration by Truth Discovery-regularized Optimization}\label{sec:exp-eces}

\paragraph{Experimental Setup.}
In this section, 
%our goal is to 
we evaluate the performance of our truth discovery-regularized post-hoc calibration methods (pTDE), to which the information elicited from ensemble-based methods is incorporated. To this end, we train ResNet18, DenseNet121 \citep{DenseNet} and ResNeXt29 \citep{ResNeXt} on CIFAR10/100 datasets with Snapshot Ensemble scheme and obtain 200 ensemble members as the sources (i.e., $S=200$). See Section \ref{supp:exp} for details. The pre-trained ImageNet models are also used with source numbers set at 10 and 30 for both DE and SE. We first train the vanilla \underline{opt}imization method using \underline{hist}ogram-based ECE as the loss function (opt\textsuperscript{\emph{hist}}) as described in Section \ref{para:opt-hist-ece} with batch size at 1000 for CIFARs and 10000 for ImageNet for 70 epochs. Then, we apply compositional training by switching to \underline{KDE}-based loss function (opt\textsuperscript{\emph{KDE}}) for 5 additional epochs. To leverage the information gained from truth discovery, the entropy based geometric variance (HV) values are computed for all ($N_c+N_e$) samples. By taking HV into the training process, opt\textsuperscript{\emph{hist}} is promoted to truth discovery-regularized \underline{p}ost-hoc calibration pTDE\textsuperscript{\emph{hist}}, which is subsequently optimized for 5 additional epochs using ECE\textsuperscript{\emph{KDE}} as the loss function to be pTDE\textsuperscript{\emph{KDE}}. Finally, pTDE\textsuperscript{\emph{KDE}} (or pTDE for short) is compared with several state-of-the-art post-hoc calibration methods, namely, Temperature Scaling (TS) \citep{guo2017calibration}, Ensemble Temperature Scaling (ETS) \citep{mixnmatch}, and multi-class isotonic regression (IRM) \citep{mixnmatch}. 

% ~~~~~~~~~~~~~~~~~~~~ ~~~~~~~~~~~~~~~~~~~~ ~~~~~~~~~~~~~~~~~~~~ table: bin plot begin
\begin{figure}[t]
  \centering
  \includegraphics[width=1.0\linewidth]{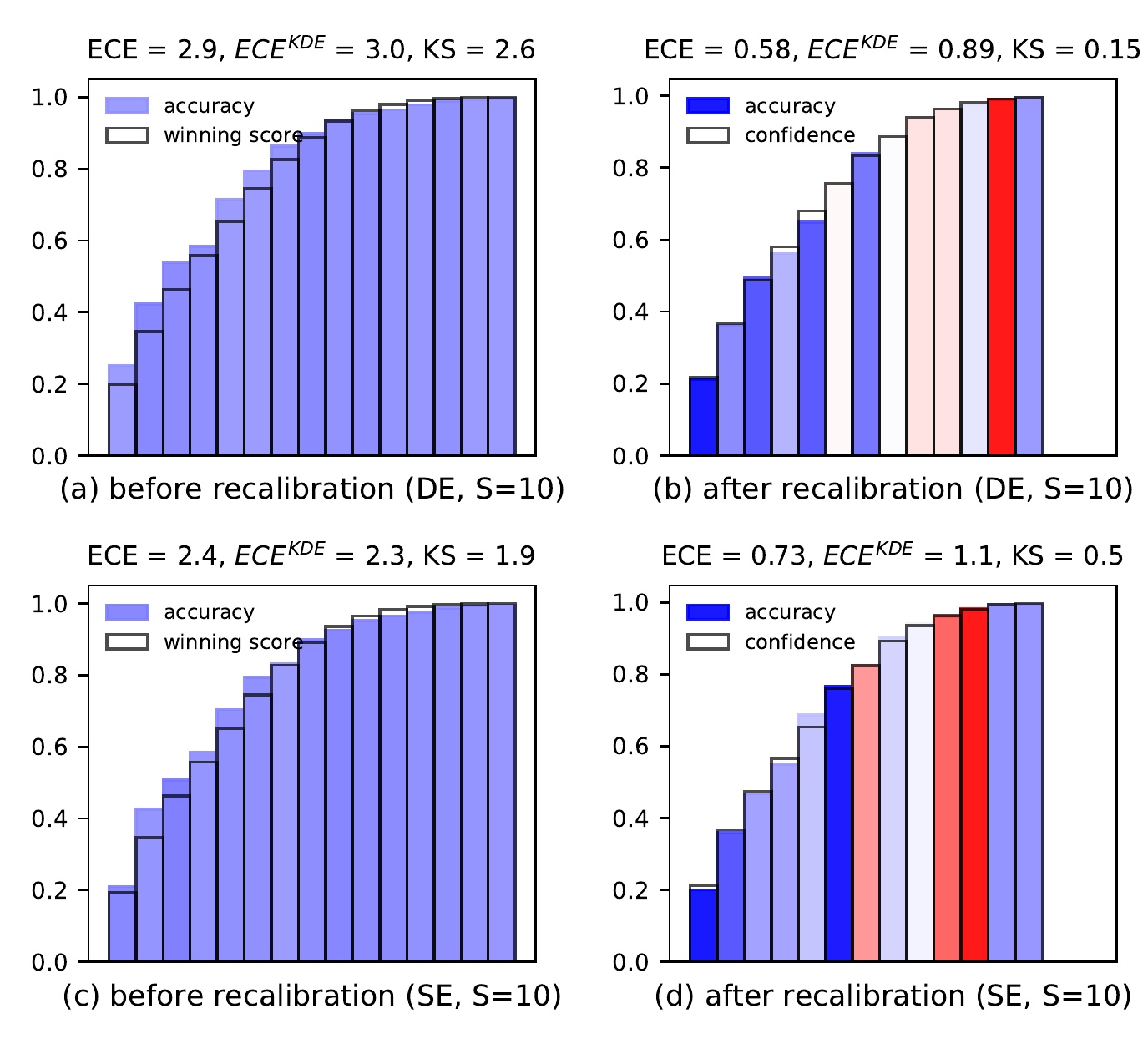}
  \caption{Winning score/confidence vs. actual accuracy on the unseen ImageNet evaluation dataset for Deep Ensemble (a, b) and Snapshot Ensemble (c, d), both with 10 sources, before (a, c) and after (b, d) the post-hoc calibration. The initial bins in (a, c) are selected such that all samples are evenly distributed over the bins, as suggested by \citep{mixnmatch}. Then we fix the positions of bins in (b, d) for a better illustration. The color temperature of each bin reflects the number of samples that fall into that bin. The warmer the color is, the more samples the bin contains, and vice versa.}\label{fig:binplot}
\end{figure}
% ~~~~~~~~~~~~~~~~~~~~ ~~~~~~~~~~~~~~~~~~~~ ~~~~~~~~~~~~~~~~~~~~ table: bin plot end

\paragraph{Results.}
The attenuation/enhancement factor we apply in Eq. (\ref{eq:factor}) enables an iterative rearrangement of samples across bins, until a small discrepancy between confidence and actual accuracy is achieved within every bin. Figure \ref{fig:binplot} shows how this recalibration (vanilla opt\textsuperscript{\emph{hist}}) affects confidence distribution. From Table \ref{tab:geoC}, we can see that our method (pTDE) significantly outperforms all competing methods (except on CIFAR10 with ECE). It is worth noting that when pTDE is excluded, there is no sweeping method under every circumstances, but pTDE overall shows better consistency especially with the ECE\textsuperscript{\emph{KDE}} metric which is not susceptible to the binning strategy, e.g., the number and positions of the bins. To inspect the individual contributions of compositional training and truth discovery regularization to the total performance, we conduct ablation study for the two components as shown in Table \ref{tab:ablation}. Generally, the truth discover-regularized methods (pTDE\textsuperscript{\emph{hist}} and pTDE\textsuperscript{\emph{KDE}}) perform better. Depending on which kind of calibration error is used as loss function, it is purposefully minimized. This indicates that  pTDE\textsuperscript{\emph{hist}} performs better with ECE metric, while  pTDE\textsuperscript{\emph{KDE}} favors ECE\textsuperscript{\emph{KDE}} metric. Notably, although ECE\textsuperscript{\emph{KDE}} is targeted by pTDE\textsuperscript{\emph{KDE}}, ECE can still be decreased on ImageNet, for example, when 30 sources from Deep Ensemble is used, showing the effectiveness of our proposed compositional training and truth discovery regularization. To get more insight into why truth discovery improves calibration performance, we show the relationship of the computed Entropy based geometric variance (HV) with the winning score in Figures \ref{supp:cifar10-densnet} \ref{supp:cifar100-resnet} \ref{supp:imagenet-dee} showing that truth discovery indeed provides information that orthogonal to winning score itself, and thus to prevent overfitting to the calibration dataset. 
%\ref{supp:imagenet-sse}

% ~~~~~~~~~~~~~~~~~~~~ ~~~~~~~~~~~~~~~~~~~~ ~~~~~~~~~~~~~~~~~~~~ table: KS begin
\begin{table}[!htp]\centering
  \caption{KS error (in \%) on ImageNet evaluation dataset by various post-hoc calibration methods including four variants of our proposed method. The best results are shown in bold.}\label{tab:ks}
  \small
  \begin{tabular}{rrr|rrr}\toprule
  &\multicolumn{2}{c|}{Deep Ensemble} &\multicolumn{2}{c}{Snapshot Ensemble} \\\cmidrule{2-5}
  Method &$S$ = 10 &$S$ = 30  &$S$ = 10  &$S$ = 30 \\\midrule
  DE/SE &2.71{\scriptsize $\pm$0.10} &2.86{\scriptsize $\pm$0.03} &1.83{\scriptsize $\pm$0.08} &1.52{\scriptsize $\pm$0.09} \\
  TS &0.88{\scriptsize $\pm$0.13} &1.03{\scriptsize $\pm$0.09} &1.02{\scriptsize $\pm$0.12} &1.12{\scriptsize $\pm$0.15} \\
  ETS &0.59{\scriptsize $\pm$0.13} &0.42{\scriptsize $\pm$0.10} &0.50{\scriptsize $\pm$0.11} &0.49{\scriptsize $\pm$0.15} \\
  IRM &0.97{\scriptsize $\pm$0.13} &0.93{\scriptsize $\pm$0.10} &0.75{\scriptsize $\pm$0.09} &0.66{\scriptsize $\pm$0.14} \\
  Spline &0.38{\scriptsize $\pm$0.09} &0.34{\scriptsize $\pm$0.07} &0.27{\scriptsize $\pm$0.08} &\textbf{0.30{\scriptsize $\pm$0.11}} \\ \midrule
  \scriptsize{opt\textsuperscript{\emph{hist}}} &\cellcolor[HTML]{f3f3f3}0.43{\scriptsize $\pm$0.20} &\cellcolor[HTML]{f3f3f3}0.39{\scriptsize $\pm$0.16} &\cellcolor[HTML]{f3f3f3}0.41{\scriptsize $\pm$0.06} &\cellcolor[HTML]{f3f3f3}0.36{\scriptsize $\pm$0.17} \\
  \scriptsize{opt\textsuperscript{\emph{KDE}}} &\cellcolor[HTML]{f3f3f3}0.37{\scriptsize $\pm$0.15} &\cellcolor[HTML]{f3f3f3}0.33{\scriptsize $\pm$0.11} &\cellcolor[HTML]{f3f3f3}0.31{\scriptsize $\pm$0.07} &\cellcolor[HTML]{f3f3f3}0.33{\scriptsize $\pm$0.14} \\
  \scriptsize{pTDE\textsuperscript{\emph{hist}}} &\cellcolor[HTML]{f3f3f3}0.42{\scriptsize $\pm$0.18} &\cellcolor[HTML]{f3f3f3}0.36{\scriptsize $\pm$0.19} &\cellcolor[HTML]{f3f3f3}0.26{\scriptsize $\pm$0.06} &\cellcolor[HTML]{f3f3f3}0.34{\scriptsize $\pm$0.18} \\
  \scriptsize{pTDE\textsuperscript{\emph{KDE}}} &\cellcolor[HTML]{f3f3f3}\textbf{0.37{\scriptsize $\pm$0.14}} &\cellcolor[HTML]{f3f3f3}\textbf{0.27{\scriptsize $\pm$0.12}} &\cellcolor[HTML]{f3f3f3}\textbf{0.25{\scriptsize $\pm$0.04}} &\cellcolor[HTML]{f3f3f3}\textbf{0.31{\scriptsize $\pm$0.13}} \\
  \bottomrule
  \end{tabular}
\end{table}

Finally, all the calibration results are further measured upon KS, shown in Table \ref{tab:ks}, and surprisingly, although pTDE is not specifically designed for the optimization of KS, its fully-fledged version pTDE\textsuperscript{\emph{KDE}} is competitive, or even better than Spline, showing that the truth discovery-based regularizer is also beneficial to the minimization of the KS metric.

% ==================== ==================== ==================== ==================== ====================
% 5 Conclusion
% ==================== ==================== ==================== ==================== ====================
\section{Conclusion}\label{sec:conc} 

In this work, we first present Truth Discovery Ensemble (TDE) that neither requires hold-out calibration data nor alters any training process, but significantly surpasses the original ensemble result, and in the meanwhile preserves the accuracy (aTDE). For post-hoc calibration, the superiority of our final methods (pTDE) is attributed not only to truth discovery, but also to the compositional training strategy. In conclusion, truth discovery is well positioned to assist both ensemble-based and post-hoc calibration. We hope that the proposed calibrators augmented by truth discovery can enlarge the arsenal of uncertainty calibration methods for deep learning. Our source code is available at https://github.com/horsepurve/truly-uncertain.
% ~~~~~~~~~~~~~~~~~~~~ ~~~~~~~~~~~~~~~~~~~~ ~~~~~~~~~~~~~~~~~~~~ table: KS end

% ==================== ==================== ==================== ==================== ====================
%                                               End of Main Text                                             %
% ==================== ==================== ==================== ==================== ====================
% \clearpage % test 

\iffalse
  \begin{contributions} % will be removed in pdf for initial submission,
                        % so you can already fill it to test with the
                        % ‘accepted’ class option
      The authors conceived the idea and wrote the paper.
  \end{contributions}
\fi

\begin{acknowledgements} % will be removed in pdf for initial submission,
                         % so you can already fill it to test with the
                         % ‘accepted’ class option
  This research was supported in part by NSF through grants CCF-1716400 and IIS-1910492.     
\end{acknowledgements}

\bibliography{ma_46}

% ==================== ==================== ==================== ==================== ====================
%                                               End of Paper                                             %
% ==================== ==================== ==================== ==================== ====================

\clearpage
\onecolumn

\appendix

\setcounter{table}{0}
\setcounter{figure}{0}
\renewcommand{\thetable}{A\arabic{table}}
\renewcommand{\thefigure}{A\arabic{figure}}

\section{Appendix}
\subsection{Proof of Theorem 3.1}\label{supp:proof}
\begin{proof}
For the updated truth vector $\mathbf{z}^*$ in each iteration, we want it to maintain the prediction accuracy. That is to say, the index of the maximum component of $\mathbf{z}^*$ could not be different from that of $\mathbf{z}_{ens}$, i.e., $\operatorname*{arg\,max}_l z_l^* = \operatorname*{arg\,max}_l (z_{ens})_l$. Let $c = \operatorname*{arg\,max}_l (z_{ens})_l$. Then,  we want to find the projection of $\mathbf{z}^*$ onto the accuracy-preserving simplex $\triangle_a: \{z_c > z_l, \forall l \neq c; 0 \leq z_{l'} \leq 1, \forall l'; \sum_{l'=1}^L z_{l'} = 1\}$. This projection can be found through the following constrained optimization problem:
\begin{equation}\label{eq:supp-1}
  \begin{problem}
    \minimize_{\{z_1,z_2,...,z_L\}} \quad& (z_1 - z_1^*)^2 + (z_2 - z_2^*)^2 + ... + (z_L - z_L^*)^2 \\
    \textrm{s.t.} \quad& z_l < z_c \quad \forall l \neq c \\
    \quad& 0 \leq z_{l'} \leq 1 \quad \forall l' \\
    \quad& \sum_{l'=1}^{L} z_{l'} = 1.
  \end{problem}
\end{equation}

We solve the above optimization problem using Lagrange Multipliers with Karush-Kuhn-Tucker (KKT) Conditions. Let $\lambda_0, \lambda_1,...,\lambda_{c-1} \geq 0, \eta_1,...,\eta_{c-1} \in \mathbb{R}$ and define
\begin{equation*}
    G = \sum_{l'=1}^{L} (z_{l'} - z_{l'}^*)^2 
    + \sum_{\forall l \neq c} \lambda_l (z_1 - z_c + \eta_l^2)
    + \lambda_0 (z_1+...+z_L-1). 
\end{equation*}

Taking the derivative with respect to $\mathbf{z}$ and the multipliers gives
\begin{align*}
    &\frac{\partial G}{\partial z_c} = 2(z_c-z_c^*)+\lambda_0+\lambda_1+...+\lambda_{c-1} \\
    &\frac{\partial G}{\partial z_l} = 2(z_l^* - z_l) + \lambda_0 - \lambda_l \quad \forall l \neq c \\
    &\frac{\partial G}{\partial \lambda_l} = z_l - z_c + \eta_l^2 \quad \forall l \neq c \\
    &\frac{\partial G}{\partial \eta_l} = 2\lambda_l\eta_l \quad \forall l \neq c. 
\end{align*}
Setting the gradient of the Lagrangian $G$ to 0 and rearranging gives us $\lambda_{l} = 2(z_1^* - z_1)$, $z_l - z_c + \eta_l^* = 0$, and $\lambda_l \eta_l = 0$, $\forall l \neq c$. For every $\lambda_l \leq 0$, there are two possibilities: (I) if $\lambda_l = 0$, then $z_l = z_l^*$ which means that the $l^{th}$ components will not be changed; (II) if $\lambda_l > 0$, then $z_l$ will be decreased to be the same as $z_c$. In Algorithm \ref{alg:acc}, we first sort the components of $\mathbf{z^*}$, denoted as $\{z_l^{*\prime}\}_{l=1}^L$, then the components less than $z_c^*$ should be fixed, or otherwise it will contradict the above two conditions. Then, we iterate through the top $\tilde{l}$ components of the sorted components until we find an $\tilde{l}$ such that the following condition holds:
\begin{equation}\label{eq:supp-check}
    \frac{1}{\tilde{l}+1}(z_c^* + \sum_{l'=1}^{\tilde{l}}z^{*\prime}_{l'}) > z^{*\prime}_{\tilde{l}+1}.    
\end{equation}
Since $\{z_l^{*\prime}\}_{l=1}^L$ is sorted,  Inequal. (\ref{eq:supp-check}) implies that all conditions in problem (\ref{eq:supp-1}) hold. Therefore, the solution 
\begin{align*}
    \tilde{z}_{l'} = 
    \begin{cases}
    \frac{1}{\tilde{l}+1}(z_c^* + \sum_{l'=1}^{\tilde{l}}z^{*\prime}_{l'}) \quad &\forall l' \leq \tilde{l} \\
    z_{l'}^{*\prime} \quad &\forall l' > \tilde{l}
    \end{cases}
\end{align*}
given by Algorithm 3.1 will not alter the index of the maximum component. %$\Box$
\end{proof}

\subsection{Experimental Details}\label{supp:exp}
In this section we describe the detailed experimental setups for post-hoc calibration. Three architectures ResNet18, DenseNet121, and ResNeXt29 are trained on CIFAR10/100 for 200 epochs with a batch size of 128. The learning rate is 0.1 initially and multiplied by 0.2 at the $60^{th}$, $120^{th}$, and $160^{th}$ epoch. For all experiment SGD optimizer is used with momentum at 0.9 and weight decay at 5e-4. To obtain as many sources as possible in reasonable time, we treat all models from 200 snapshots as sources for truth discovery and for the computation of the Entropy based Geometric Variance (HV). In the post-hoc calibration step, all the original testing sets are further randomly split into two equal-sized calibration/evaluation datasets, i.e., $N_c=N_e=5000$ for CIFAR10/100 and $N_c=N_e=25000$ for ImageNet. This random split is performed 5 times using different seeds, and the resulting means and standard deviations are reported. For all the experiments, the number of bins $B$ is set at 15 and their endpoints are determined by evenly distributing all calibration samples into the 15 bins. Then the number and locations of the bins are fixed during the optimizations. As for the hyperparameters in Eq. (\ref{eq:factor-hv}), we simply set $\alpha_1 = 1, \alpha_2 = \psi_{\kappa}$ without manually tuning, in order to keep minimal human intervention.

% ==================== ==================== ==================== ==================== ====================
%                                        Additional Tables & Figures                                     %
% ==================== ==================== ==================== ==================== ====================
\subsection{Additional Tables \& Figures}\label{supp:figtab}
% ==================== ==================== PreResNet110 ==================== ====================

\begin{table}[!htp]\centering
  \caption{Comparison between DE/TDE/aTDE on PreResNet110. The best results are highlighted.}\label{tab:supp-truth1}
  \scriptsize
  \resizebox{\textwidth}{!}{%
  \begin{tabular}{rc|rrrrrr|rrrrrrr}\toprule
  \textbf{} &\textbf{} &\multicolumn{6}{c}{\textbf{CIFAR100}} &\multicolumn{6}{c}{\textbf{CIFAR10}} \\\cmidrule{3-14}
  \textbf{} &\textbf{\#sources} &\textbf{ECE\textsuperscript{\emph{KDE}}$\downarrow$} &\textbf{ECE$\downarrow$} &\textbf{NLL$\downarrow$} &\textbf{MSE$\downarrow$} &\textbf{KS$\downarrow$} &\textbf{ACC$\uparrow$} &\textbf{ECE\textsuperscript{\emph{KDE}}$\downarrow$} &\textbf{ECE$\downarrow$} &\textbf{NLL$\downarrow$} &\textbf{MSE$\downarrow$} &\textbf{KS$\downarrow$} &\textbf{ACC$\uparrow$} \\\midrule
  \textbf{DE} &\multirow{3}{*}{10} &\ul{2.22} &2.10 &\ul{0.638} &\ul{0.252} &\ul{1.61} &\ul{82.32} &\ul{1.10} &\ul{0.50} &\ul{0.113} &\ul{0.054} &\ul{0.19} &96.28 \\
  \textbf{TDE} & &\cellcolor[HTML]{f3f3f3}2.27 &\cellcolor[HTML]{f3f3f3}3.06 &\cellcolor[HTML]{f3f3f3}0.648 &\cellcolor[HTML]{f3f3f3}0.254 &\cellcolor[HTML]{f3f3f3}2.78 &\cellcolor[HTML]{f3f3f3}82.21 &\cellcolor[HTML]{f3f3f3}1.38 &\cellcolor[HTML]{f3f3f3}1.35 &\cellcolor[HTML]{f3f3f3}0.119 &\cellcolor[HTML]{f3f3f3}0.056 &\cellcolor[HTML]{f3f3f3}1.35 &\cellcolor[HTML]{f3f3f3}\ul{96.32} \\
  \textbf{aTDE} & &\cellcolor[HTML]{f3f3f3}2.37 &\cellcolor[HTML]{f3f3f3}3.16 &\cellcolor[HTML]{f3f3f3}0.648 &\cellcolor[HTML]{f3f3f3}0.254 &\cellcolor[HTML]{f3f3f3}2.63 &\cellcolor[HTML]{f3f3f3}\ul{82.32} &\cellcolor[HTML]{f3f3f3}1.43 &\cellcolor[HTML]{f3f3f3}1.38 &\cellcolor[HTML]{f3f3f3}0.119 &\cellcolor[HTML]{f3f3f3}0.056 &\cellcolor[HTML]{f3f3f3}1.38 &\cellcolor[HTML]{f3f3f3}96.28 \\ \midrule
  \textbf{DE} &\multirow{3}{*}{20} &2.65 &2.43 &\ul{0.618} &0.247 &2.07 &\ul{82.67} &1.25 &\ul{0.45} &\ul{0.110} &\ul{0.053} &\ul{0.23} &\ul{96.42} \\
  \textbf{TDE} & &\cellcolor[HTML]{f3f3f3}1.68 &\cellcolor[HTML]{f3f3f3}\ul{1.90} &\cellcolor[HTML]{f3f3f3}0.621 &\cellcolor[HTML]{f3f3f3}0.247 &\cellcolor[HTML]{f3f3f3}1.48 &\cellcolor[HTML]{f3f3f3}82.61 &\cellcolor[HTML]{f3f3f3}1.23 &\cellcolor[HTML]{f3f3f3}1.00 &\cellcolor[HTML]{f3f3f3}0.112 &\cellcolor[HTML]{f3f3f3}0.054 &\cellcolor[HTML]{f3f3f3}0.99 &\cellcolor[HTML]{f3f3f3}96.39 \\
  \textbf{aTDE} & &\cellcolor[HTML]{f3f3f3}\ul{1.66} &\cellcolor[HTML]{f3f3f3}1.94 &\cellcolor[HTML]{f3f3f3}0.620 &\cellcolor[HTML]{f3f3f3}\ul{0.247} &\cellcolor[HTML]{f3f3f3}\ul{1.41} &\cellcolor[HTML]{f3f3f3}\ul{82.67} &\cellcolor[HTML]{f3f3f3}\ul{1.21} &\cellcolor[HTML]{f3f3f3}0.97 &\cellcolor[HTML]{f3f3f3}0.112 &\cellcolor[HTML]{f3f3f3}0.054 &\cellcolor[HTML]{f3f3f3}0.96 &\cellcolor[HTML]{f3f3f3}\ul{96.42} \\ \midrule
  \textbf{DE} &\multirow{3}{*}{30} &2.93 &2.69 &\ul{0.610} &0.245 &2.33 &\ul{82.92} &1.27 &\ul{0.43} &\ul{0.107} &\ul{0.052} &\ul{0.25} &\ul{96.42} \\
  \textbf{TDE} & &\cellcolor[HTML]{f3f3f3}\ul{1.59} &\cellcolor[HTML]{f3f3f3}\ul{1.92} &\cellcolor[HTML]{f3f3f3}0.612 &\cellcolor[HTML]{f3f3f3}0.245 &\cellcolor[HTML]{f3f3f3}0.85 &\cellcolor[HTML]{f3f3f3}82.87 &\cellcolor[HTML]{f3f3f3}\ul{1.15} &\cellcolor[HTML]{f3f3f3}0.83 &\cellcolor[HTML]{f3f3f3}0.109 &\cellcolor[HTML]{f3f3f3}0.053 &\cellcolor[HTML]{f3f3f3}0.82 &\cellcolor[HTML]{f3f3f3}\ul{96.42} \\
  \textbf{aTDE} & &\cellcolor[HTML]{f3f3f3}1.61 &\cellcolor[HTML]{f3f3f3}1.97 &\cellcolor[HTML]{f3f3f3}0.612 &\cellcolor[HTML]{f3f3f3}\ul{0.245} &\cellcolor[HTML]{f3f3f3}\ul{0.80} &\cellcolor[HTML]{f3f3f3}\ul{82.92} &\cellcolor[HTML]{f3f3f3}1.19 &\cellcolor[HTML]{f3f3f3}0.83 &\cellcolor[HTML]{f3f3f3}0.109 &\cellcolor[HTML]{f3f3f3}0.053 &\cellcolor[HTML]{f3f3f3}0.82 &\cellcolor[HTML]{f3f3f3}\ul{96.42} \\ \midrule
  \textbf{DE} &\multirow{3}{*}{40} &2.83 &2.48 &\ul{0.606} &0.244 &2.17 &82.79 &1.33 &\ul{0.41} &\ul{0.107} &\ul{0.052} &\ul{0.25} &\ul{96.41} \\
  \textbf{TDE} & &\cellcolor[HTML]{f3f3f3}1.61 &\cellcolor[HTML]{f3f3f3}2.05 &\cellcolor[HTML]{f3f3f3}0.606 &\cellcolor[HTML]{f3f3f3}\ul{0.244} &\cellcolor[HTML]{f3f3f3}0.72 &\cellcolor[HTML]{f3f3f3}\ul{82.86} &\cellcolor[HTML]{f3f3f3}\ul{1.09} &\cellcolor[HTML]{f3f3f3}0.78 &\cellcolor[HTML]{f3f3f3}0.109 &\cellcolor[HTML]{f3f3f3}\ul{0.052} &\cellcolor[HTML]{f3f3f3}0.77 &\cellcolor[HTML]{f3f3f3}96.39 \\
  \textbf{aTDE} & &\cellcolor[HTML]{f3f3f3}\ul{1.54} &\cellcolor[HTML]{f3f3f3}\ul{1.98} &\cellcolor[HTML]{f3f3f3}0.606 &\cellcolor[HTML]{f3f3f3}0.244 &\cellcolor[HTML]{f3f3f3}\ul{0.71} &\cellcolor[HTML]{f3f3f3}82.79 &\cellcolor[HTML]{f3f3f3}\ul{1.09} &\cellcolor[HTML]{f3f3f3}0.76 &\cellcolor[HTML]{f3f3f3}0.109 &\cellcolor[HTML]{f3f3f3}\ul{0.052} &\cellcolor[HTML]{f3f3f3}0.75 &\cellcolor[HTML]{f3f3f3}\ul{96.41} \\ \midrule
  \textbf{DE} &\multirow{3}{*}{50} &2.89 &2.50 &\ul{0.602} &0.244 &2.20 &82.83 &1.31 &\ul{0.48} &\ul{0.106} &\ul{0.052} &\ul{0.23} &\ul{96.38} \\
  \textbf{TDE} & &\cellcolor[HTML]{f3f3f3}1.61 &\cellcolor[HTML]{f3f3f3}1.99 &\cellcolor[HTML]{f3f3f3}0.602 &\cellcolor[HTML]{f3f3f3}\ul{0.243} &\cellcolor[HTML]{f3f3f3}0.78 &\cellcolor[HTML]{f3f3f3}\ul{82.89} &\cellcolor[HTML]{f3f3f3}\ul{1.06} &\cellcolor[HTML]{f3f3f3}0.76 &\cellcolor[HTML]{f3f3f3}0.107 &\cellcolor[HTML]{f3f3f3}\ul{0.052} &\cellcolor[HTML]{f3f3f3}0.75 &\cellcolor[HTML]{f3f3f3}\ul{96.38} \\
  \textbf{aTDE} & &\cellcolor[HTML]{f3f3f3}\ul{1.55} &\cellcolor[HTML]{f3f3f3}\ul{1.93} &\cellcolor[HTML]{f3f3f3}0.602 &\cellcolor[HTML]{f3f3f3}0.243 &\cellcolor[HTML]{f3f3f3}\ul{0.72} &\cellcolor[HTML]{f3f3f3}82.83 &\cellcolor[HTML]{f3f3f3}1.07 &\cellcolor[HTML]{f3f3f3}0.76 &\cellcolor[HTML]{f3f3f3}0.107 &\cellcolor[HTML]{f3f3f3}\ul{0.052} &\cellcolor[HTML]{f3f3f3}0.75 &\cellcolor[HTML]{f3f3f3}\ul{96.38} \\ \midrule
  \textbf{DE} &\multirow{3}{*}{60} &3.06 &2.62 &\ul{0.600} &0.243 &2.39 &\ul{83.00} &1.31 &\ul{0.38} &\ul{0.106} &\ul{0.052} &\ul{0.23} &\ul{96.45} \\
  \textbf{TDE} & &\cellcolor[HTML]{f3f3f3}\ul{1.75} &\cellcolor[HTML]{f3f3f3}\ul{1.98} &\cellcolor[HTML]{f3f3f3}0.600 &\cellcolor[HTML]{f3f3f3}\ul{0.242} &\cellcolor[HTML]{f3f3f3}\ul{0.92} &\cellcolor[HTML]{f3f3f3}82.96 &\cellcolor[HTML]{f3f3f3}\ul{1.05} &\cellcolor[HTML]{f3f3f3}0.71 &\cellcolor[HTML]{f3f3f3}0.107 &\cellcolor[HTML]{f3f3f3}\ul{0.052} &\cellcolor[HTML]{f3f3f3}0.69 &\cellcolor[HTML]{f3f3f3}96.40 \\
  \textbf{aTDE} & &\cellcolor[HTML]{f3f3f3}1.79 &\cellcolor[HTML]{f3f3f3}2.02 &\cellcolor[HTML]{f3f3f3}0.600 &\cellcolor[HTML]{f3f3f3}0.242 &\cellcolor[HTML]{f3f3f3}0.96 &\cellcolor[HTML]{f3f3f3}\ul{83.00} &\cellcolor[HTML]{f3f3f3}1.09 &\cellcolor[HTML]{f3f3f3}0.66 &\cellcolor[HTML]{f3f3f3}0.107 &\cellcolor[HTML]{f3f3f3}\ul{0.052} &\cellcolor[HTML]{f3f3f3}0.64 &\cellcolor[HTML]{f3f3f3}\ul{96.45} \\ \midrule
  \textbf{DE} &\multirow{3}{*}{70} &3.08 &2.63 &\ul{0.598} &0.243 &2.41 &83.02 &1.24 &\ul{0.39} &\ul{0.106} &\ul{0.052} &\ul{0.21} &\ul{96.42} \\
  \textbf{TDE} & &\cellcolor[HTML]{f3f3f3}1.85 &\cellcolor[HTML]{f3f3f3}2.02 &\cellcolor[HTML]{f3f3f3}0.599 &\cellcolor[HTML]{f3f3f3}\ul{0.242} &\cellcolor[HTML]{f3f3f3}1.03 &\cellcolor[HTML]{f3f3f3}\ul{83.04} &\cellcolor[HTML]{f3f3f3}\ul{1.04} &\cellcolor[HTML]{f3f3f3}0.70 &\cellcolor[HTML]{f3f3f3}0.107 &\cellcolor[HTML]{f3f3f3}\ul{0.052} &\cellcolor[HTML]{f3f3f3}0.68 &\cellcolor[HTML]{f3f3f3}96.38 \\
  \textbf{aTDE} & &\cellcolor[HTML]{f3f3f3}\ul{1.82} &\cellcolor[HTML]{f3f3f3}\ul{2.00} &\cellcolor[HTML]{f3f3f3}0.599 &\cellcolor[HTML]{f3f3f3}\ul{0.242} &\cellcolor[HTML]{f3f3f3}\ul{1.01} &\cellcolor[HTML]{f3f3f3}83.02 &\cellcolor[HTML]{f3f3f3}\ul{1.04} &\cellcolor[HTML]{f3f3f3}0.66 &\cellcolor[HTML]{f3f3f3}0.107 &\cellcolor[HTML]{f3f3f3}\ul{0.052} &\cellcolor[HTML]{f3f3f3}0.64 &\cellcolor[HTML]{f3f3f3}\ul{96.42} \\ \midrule
  \textbf{DE} &\multirow{3}{*}{80} &3.06 &2.63 &\ul{0.596} &0.243 &2.38 &\ul{83.00} &1.21 &\ul{0.39} &\ul{0.105} &\ul{0.052} &\ul{0.25} &\ul{96.46} \\
  \textbf{TDE} & &\cellcolor[HTML]{f3f3f3}\ul{1.76} &\cellcolor[HTML]{f3f3f3}\ul{2.00} &\cellcolor[HTML]{f3f3f3}0.597 &\cellcolor[HTML]{f3f3f3}\ul{0.242} &\cellcolor[HTML]{f3f3f3}\ul{1.01} &\cellcolor[HTML]{f3f3f3}\ul{83.00} &\cellcolor[HTML]{f3f3f3}1.03 &\cellcolor[HTML]{f3f3f3}0.59 &\cellcolor[HTML]{f3f3f3}0.106 &\cellcolor[HTML]{f3f3f3}\ul{0.052} &\cellcolor[HTML]{f3f3f3}0.59 &\cellcolor[HTML]{f3f3f3}96.45 \\
  \textbf{aTDE} & &\cellcolor[HTML]{f3f3f3}1.78 &\cellcolor[HTML]{f3f3f3}\ul{2.00} &\cellcolor[HTML]{f3f3f3}0.597 &\cellcolor[HTML]{f3f3f3}\ul{0.242} &\cellcolor[HTML]{f3f3f3}\ul{1.01} &\cellcolor[HTML]{f3f3f3}\ul{83.00} &\cellcolor[HTML]{f3f3f3}\ul{1.01} &\cellcolor[HTML]{f3f3f3}0.58 &\cellcolor[HTML]{f3f3f3}0.106 &\cellcolor[HTML]{f3f3f3}\ul{0.052} &\cellcolor[HTML]{f3f3f3}0.58 &\cellcolor[HTML]{f3f3f3}\ul{96.46} \\ \midrule
  \textbf{DE} &\multirow{3}{*}{90} &3.06 &2.59 &\ul{0.596} &0.243 &2.39 &83.03 &1.18 &\ul{0.39} &\ul{0.105} &\ul{0.052} &\ul{0.22} &96.44 \\
  \textbf{TDE} & &\cellcolor[HTML]{f3f3f3}1.83 &\cellcolor[HTML]{f3f3f3}2.03 &\cellcolor[HTML]{f3f3f3}0.596 &\cellcolor[HTML]{f3f3f3}\ul{0.242} &\cellcolor[HTML]{f3f3f3}1.10 &\cellcolor[HTML]{f3f3f3}\ul{83.06} &\cellcolor[HTML]{f3f3f3}0.99 &\cellcolor[HTML]{f3f3f3}0.57 &\cellcolor[HTML]{f3f3f3}0.106 &\cellcolor[HTML]{f3f3f3}\ul{0.052} &\cellcolor[HTML]{f3f3f3}0.57 &\cellcolor[HTML]{f3f3f3}\ul{96.46} \\
  \textbf{aTDE} & &\cellcolor[HTML]{f3f3f3}\ul{1.80} &\cellcolor[HTML]{f3f3f3}\ul{2.00} &\cellcolor[HTML]{f3f3f3}0.596 &\cellcolor[HTML]{f3f3f3}0.242 &\cellcolor[HTML]{f3f3f3}\ul{1.07} &\cellcolor[HTML]{f3f3f3}83.03 &\cellcolor[HTML]{f3f3f3}\ul{0.98} &\cellcolor[HTML]{f3f3f3}0.59 &\cellcolor[HTML]{f3f3f3}0.106 &\cellcolor[HTML]{f3f3f3}\ul{0.052} &\cellcolor[HTML]{f3f3f3}0.59 &\cellcolor[HTML]{f3f3f3}96.44 \\ \midrule
  \textbf{DE} &\multirow{3}{*}{100} &3.07 &2.58 &\ul{0.595} &0.242 &2.38 &\ul{83.02} &1.15 &\ul{0.44} &\ul{0.106} &\ul{0.052} &\ul{0.19} &96.41 \\
  \textbf{TDE} & &\cellcolor[HTML]{f3f3f3}1.81 &\cellcolor[HTML]{f3f3f3}\ul{1.95} &\cellcolor[HTML]{f3f3f3}\ul{0.595} &\cellcolor[HTML]{f3f3f3}\ul{0.241} &\cellcolor[HTML]{f3f3f3}\ul{1.05} &\cellcolor[HTML]{f3f3f3}83.01 &\cellcolor[HTML]{f3f3f3}\ul{1.02} &\cellcolor[HTML]{f3f3f3}0.60 &\cellcolor[HTML]{f3f3f3}\ul{0.106} &\cellcolor[HTML]{f3f3f3}\ul{0.052} &\cellcolor[HTML]{f3f3f3}0.60 &\cellcolor[HTML]{f3f3f3}\ul{96.42} \\
  \textbf{aTDE} & &\cellcolor[HTML]{f3f3f3}\ul{1.78} &\cellcolor[HTML]{f3f3f3}1.96 &\cellcolor[HTML]{f3f3f3}\ul{0.595} &\cellcolor[HTML]{f3f3f3}\ul{0.241} &\cellcolor[HTML]{f3f3f3}1.06 &\cellcolor[HTML]{f3f3f3}\ul{83.02} &\cellcolor[HTML]{f3f3f3}1.03 &\cellcolor[HTML]{f3f3f3}0.61 &\cellcolor[HTML]{f3f3f3}\ul{0.106} &\cellcolor[HTML]{f3f3f3}\ul{0.052} &\cellcolor[HTML]{f3f3f3}0.61 &\cellcolor[HTML]{f3f3f3}96.41 \\
  \bottomrule
  \end{tabular}}
\end{table}
 
% ==================== ==================== PreResNet164 ==================== ====================
\begin{table}[!htp]\centering
  \caption{Comparison between DE/TDE/aTDE on PreResNet164. The best results are highlighted.}\label{tab:supp-truth2}
  \scriptsize
  \resizebox{\textwidth}{!}{%
  \begin{tabular}{rc|rrrrrr|rrrrrrr}\toprule
  \textbf{} &\textbf{} &\multicolumn{6}{c}{\textbf{CIFAR100}} &\multicolumn{6}{c}{\textbf{CIFAR10}} \\\cmidrule{3-14}
  \textbf{} &\textbf{\#sources} &\textbf{ECE\textsuperscript{\emph{KDE}}$\downarrow$} &\textbf{ECE$\downarrow$} &\textbf{NLL$\downarrow$} &\textbf{MSE$\downarrow$} &\textbf{KS$\downarrow$} &\textbf{ACC$\uparrow$} &\textbf{ECE\textsuperscript{\emph{KDE}}$\downarrow$} &\textbf{ECE$\downarrow$} &\textbf{NLL$\downarrow$} &\textbf{MSE$\downarrow$} &\textbf{KS$\downarrow$} &\textbf{ACC$\uparrow$} \\\midrule
  \textbf{DE} &\multirow{3}{*}{10} &\ul{1.90} &\ul{1.86} &\ul{0.618} &\ul{0.242} &\ul{1.00} &82.95 &\ul{0.93} &\ul{0.32} &\ul{0.108} &\ul{0.052} &\ul{0.17} &\ul{96.57} \\
  \textbf{TDE} & &\cellcolor[HTML]{f3f3f3}2.66 &\cellcolor[HTML]{f3f3f3}3.27 &\cellcolor[HTML]{f3f3f3}0.632 &\cellcolor[HTML]{f3f3f3}0.245 &\cellcolor[HTML]{f3f3f3}3.14 &\cellcolor[HTML]{f3f3f3}\ul{83.02} &\cellcolor[HTML]{f3f3f3}1.25 &\cellcolor[HTML]{f3f3f3}1.25 &\cellcolor[HTML]{f3f3f3}0.115 &\cellcolor[HTML]{f3f3f3}0.053 &\cellcolor[HTML]{f3f3f3}1.24 &\cellcolor[HTML]{f3f3f3}96.55 \\
  \textbf{aTDE} & &\cellcolor[HTML]{f3f3f3}2.60 &\cellcolor[HTML]{f3f3f3}3.18 &\cellcolor[HTML]{f3f3f3}0.632 &\cellcolor[HTML]{f3f3f3}0.245 &\cellcolor[HTML]{f3f3f3}3.18 &\cellcolor[HTML]{f3f3f3}82.95 &\cellcolor[HTML]{f3f3f3}1.28 &\cellcolor[HTML]{f3f3f3}1.22 &\cellcolor[HTML]{f3f3f3}0.115 &\cellcolor[HTML]{f3f3f3}0.053 &\cellcolor[HTML]{f3f3f3}1.22 &\cellcolor[HTML]{f3f3f3}\ul{96.57} \\ \midrule
  \textbf{DE} &\multirow{3}{*}{20} &2.05 &\ul{1.87} &\ul{0.595} &\ul{0.238} &\ul{1.49} &\ul{83.30} &\ul{1.14} &\ul{0.28} &\ul{0.105} &\ul{0.051} &\ul{0.12} &\ul{96.54} \\
  \textbf{TDE} & &\cellcolor[HTML]{f3f3f3}1.73 &\cellcolor[HTML]{f3f3f3}2.45 &\cellcolor[HTML]{f3f3f3}0.602 &\cellcolor[HTML]{efefef}0.239 &\cellcolor[HTML]{efefef}1.86 &\cellcolor[HTML]{efefef}\ul{83.30} &\cellcolor[HTML]{efefef}1.24 &\cellcolor[HTML]{efefef}1.06 &\cellcolor[HTML]{efefef}0.109 &\cellcolor[HTML]{efefef}0.052 &\cellcolor[HTML]{efefef}1.06 &\cellcolor[HTML]{efefef}96.49 \\
  \textbf{aTDE} & &\cellcolor[HTML]{f3f3f3}\ul{1.67} &\cellcolor[HTML]{f3f3f3}2.46 &\cellcolor[HTML]{f3f3f3}0.602 &\cellcolor[HTML]{efefef}0.239 &\cellcolor[HTML]{efefef}1.86 &\cellcolor[HTML]{efefef}\ul{83.30} &\cellcolor[HTML]{efefef}1.24 &\cellcolor[HTML]{efefef}1.01 &\cellcolor[HTML]{efefef}0.109 &\cellcolor[HTML]{efefef}0.052 &\cellcolor[HTML]{efefef}1.01 &\cellcolor[HTML]{efefef}\ul{96.54} \\ \midrule
  \textbf{DE} &\multirow{3}{*}{30} &2.33 &2.03 &\ul{0.588} &\ul{0.236} &1.75 &\ul{83.49} &1.16 &\ul{0.26} &\ul{0.103} &\ul{0.050} &\ul{0.22} &\ul{96.62} \\
  \textbf{TDE} & &\cellcolor[HTML]{f3f3f3}\ul{1.46} &\cellcolor[HTML]{f3f3f3}\ul{2.01} &\cellcolor[HTML]{f3f3f3}0.593 &\cellcolor[HTML]{f3f3f3}0.237 &\cellcolor[HTML]{f3f3f3}1.27 &\cellcolor[HTML]{f3f3f3}83.47 &\cellcolor[HTML]{f3f3f3}\ul{1.11} &\cellcolor[HTML]{f3f3f3}0.87 &\cellcolor[HTML]{f3f3f3}0.105 &\cellcolor[HTML]{f3f3f3}\ul{0.050} &\cellcolor[HTML]{f3f3f3}0.86 &\cellcolor[HTML]{f3f3f3}96.56 \\
  \textbf{aTDE} & &\cellcolor[HTML]{f3f3f3}1.47 &\cellcolor[HTML]{f3f3f3}2.03 &\cellcolor[HTML]{f3f3f3}0.593 &\cellcolor[HTML]{f3f3f3}0.237 &\cellcolor[HTML]{f3f3f3}\ul{1.25} &\cellcolor[HTML]{f3f3f3}\ul{83.49} &\cellcolor[HTML]{f3f3f3}1.14 &\cellcolor[HTML]{f3f3f3}0.80 &\cellcolor[HTML]{f3f3f3}0.105 &\cellcolor[HTML]{f3f3f3}\ul{0.050} &\cellcolor[HTML]{f3f3f3}0.80 &\cellcolor[HTML]{f3f3f3}\ul{96.62} \\ \midrule
  \textbf{DE} &\multirow{3}{*}{40} &2.50 &2.24 &\ul{0.585} &\ul{0.236} &1.87 &\ul{83.61} &1.27 &\ul{0.31} &\ul{0.102} &\ul{0.050} &\ul{0.25} &\ul{96.66} \\
  \textbf{TDE} & &\cellcolor[HTML]{f3f3f3}\ul{1.49} &\cellcolor[HTML]{f3f3f3}\ul{1.90} &\cellcolor[HTML]{f3f3f3}0.589 &\cellcolor[HTML]{f3f3f3}\ul{0.236} &\cellcolor[HTML]{f3f3f3}0.95 &\cellcolor[HTML]{f3f3f3}83.58 &\cellcolor[HTML]{f3f3f3}\ul{1.03} &\cellcolor[HTML]{f3f3f3}0.72 &\cellcolor[HTML]{f3f3f3}0.104 &\cellcolor[HTML]{f3f3f3}\ul{0.050} &\cellcolor[HTML]{f3f3f3}0.71 &\cellcolor[HTML]{f3f3f3}96.64 \\
  \textbf{aTDE} & &\cellcolor[HTML]{f3f3f3}1.52 &\cellcolor[HTML]{f3f3f3}1.93 &\cellcolor[HTML]{f3f3f3}0.589 &\cellcolor[HTML]{f3f3f3}\ul{0.236} &\cellcolor[HTML]{f3f3f3}\ul{0.92} &\cellcolor[HTML]{f3f3f3}\ul{83.61} &\cellcolor[HTML]{f3f3f3}1.07 &\cellcolor[HTML]{f3f3f3}0.70 &\cellcolor[HTML]{f3f3f3}0.104 &\cellcolor[HTML]{f3f3f3}\ul{0.050} &\cellcolor[HTML]{f3f3f3}0.69 &\cellcolor[HTML]{f3f3f3}\ul{96.66} \\ \midrule
  \textbf{DE} &\multirow{3}{*}{50} &2.39 &2.09 &\ul{0.583} &\ul{0.235} &1.81 &\ul{83.52} &1.33 &\ul{0.35} &\ul{0.102} &\ul{0.050} &\ul{0.28} &\ul{96.68} \\
  \textbf{TDE} & &\cellcolor[HTML]{f3f3f3}\ul{1.32} &\cellcolor[HTML]{f3f3f3}\ul{1.73} &\cellcolor[HTML]{f3f3f3}0.586 &\cellcolor[HTML]{f3f3f3}\ul{0.235} &\cellcolor[HTML]{f3f3f3}0.90 &\cellcolor[HTML]{f3f3f3}83.49 &\cellcolor[HTML]{f3f3f3}\ul{1.14} &\cellcolor[HTML]{f3f3f3}0.65 &\cellcolor[HTML]{f3f3f3}0.104 &\cellcolor[HTML]{f3f3f3}\ul{0.050} &\cellcolor[HTML]{f3f3f3}0.65 &\cellcolor[HTML]{f3f3f3}96.66 \\
  \textbf{aTDE} & &\cellcolor[HTML]{f3f3f3}1.34 &\cellcolor[HTML]{f3f3f3}1.76 &\cellcolor[HTML]{f3f3f3}0.586 &\cellcolor[HTML]{f3f3f3}\ul{0.235} &\cellcolor[HTML]{f3f3f3}\ul{0.86} &\cellcolor[HTML]{f3f3f3}\ul{83.52} &\cellcolor[HTML]{f3f3f3}\ul{1.14} &\cellcolor[HTML]{f3f3f3}0.63 &\cellcolor[HTML]{f3f3f3}0.104 &\cellcolor[HTML]{f3f3f3}\ul{0.050} &\cellcolor[HTML]{f3f3f3}0.63 &\cellcolor[HTML]{f3f3f3}\ul{96.68} \\ \midrule
  \textbf{DE} &\multirow{3}{*}{60} &2.56 &2.16 &\ul{0.580} &\ul{0.234} &1.86 &\ul{83.58} &1.25 &\ul{0.31} &\ul{0.101} &\ul{0.049} &\ul{0.24} &\ul{96.66} \\
  \textbf{TDE} & &\cellcolor[HTML]{f3f3f3}\ul{1.59} &\cellcolor[HTML]{f3f3f3}\ul{1.75} &\cellcolor[HTML]{f3f3f3}0.584 &\cellcolor[HTML]{f3f3f3}\ul{0.234} &\cellcolor[HTML]{f3f3f3}0.70 &\cellcolor[HTML]{f3f3f3}83.57 &\cellcolor[HTML]{f3f3f3}\ul{1.10} &\cellcolor[HTML]{f3f3f3}0.64 &\cellcolor[HTML]{f3f3f3}0.103 &\cellcolor[HTML]{f3f3f3}0.050 &\cellcolor[HTML]{f3f3f3}0.64 &\cellcolor[HTML]{f3f3f3}96.65 \\
  \textbf{aTDE} & &\cellcolor[HTML]{f3f3f3}\ul{1.59} &\cellcolor[HTML]{f3f3f3}1.76 &\cellcolor[HTML]{f3f3f3}0.584 &\cellcolor[HTML]{f3f3f3}\ul{0.234} &\cellcolor[HTML]{f3f3f3}\ul{0.69} &\cellcolor[HTML]{f3f3f3}\ul{83.58} &\cellcolor[HTML]{f3f3f3}1.11 &\cellcolor[HTML]{f3f3f3}0.63 &\cellcolor[HTML]{f3f3f3}0.103 &\cellcolor[HTML]{f3f3f3}0.050 &\cellcolor[HTML]{f3f3f3}0.63 &\cellcolor[HTML]{f3f3f3}\ul{96.66} \\ \midrule
  \textbf{DE} &\multirow{3}{*}{70} &2.56 &2.17 &\ul{0.579} &\ul{0.234} &1.90 &83.64 &1.23 &\ul{0.27} &\ul{0.101} &\ul{0.049} &\ul{0.25} &\ul{96.66} \\
  \textbf{TDE} & &\cellcolor[HTML]{f3f3f3}\ul{1.48} &\cellcolor[HTML]{f3f3f3}1.81 &\cellcolor[HTML]{f3f3f3}0.582 &\cellcolor[HTML]{f3f3f3}\ul{0.234} &\cellcolor[HTML]{f3f3f3}0.75 &\cellcolor[HTML]{f3f3f3}\ul{83.67} &\cellcolor[HTML]{f3f3f3}\ul{1.04} &\cellcolor[HTML]{f3f3f3}0.62 &\cellcolor[HTML]{f3f3f3}0.102 &\cellcolor[HTML]{f3f3f3}\ul{0.049} &\cellcolor[HTML]{f3f3f3}0.62 &\cellcolor[HTML]{f3f3f3}96.63 \\
  \textbf{aTDE} & &\cellcolor[HTML]{f3f3f3}1.49 &\cellcolor[HTML]{f3f3f3}\ul{1.78} &\cellcolor[HTML]{f3f3f3}0.582 &\cellcolor[HTML]{f3f3f3}\ul{0.234} &\cellcolor[HTML]{f3f3f3}\ul{0.72} &\cellcolor[HTML]{f3f3f3}83.64 &\cellcolor[HTML]{f3f3f3}1.05 &\cellcolor[HTML]{f3f3f3}0.59 &\cellcolor[HTML]{f3f3f3}0.102 &\cellcolor[HTML]{f3f3f3}\ul{0.049} &\cellcolor[HTML]{f3f3f3}0.59 &\cellcolor[HTML]{f3f3f3}\ul{96.66} \\ \midrule
  \textbf{DE} &\multirow{3}{*}{80} &2.52 &2.03 &\ul{0.577} &0.234 &1.83 &\ul{83.56} &1.21 &\ul{0.25} &\ul{0.101} &\ul{0.049} &\ul{0.19} &96.59 \\
  \textbf{TDE} & &\cellcolor[HTML]{f3f3f3}\ul{1.49} &\cellcolor[HTML]{f3f3f3}\ul{1.67} &\cellcolor[HTML]{f3f3f3}0.581 &\cellcolor[HTML]{f3f3f3}\ul{0.233} &\cellcolor[HTML]{f3f3f3}\ul{0.63} &\cellcolor[HTML]{f3f3f3}83.54 &\cellcolor[HTML]{f3f3f3}\ul{1.07} &\cellcolor[HTML]{f3f3f3}0.61 &\cellcolor[HTML]{f3f3f3}0.102 &\cellcolor[HTML]{f3f3f3}\ul{0.049} &\cellcolor[HTML]{f3f3f3}0.61 &\cellcolor[HTML]{f3f3f3}\ul{96.61} \\
  \textbf{aTDE} & &\cellcolor[HTML]{f3f3f3}1.50 &\cellcolor[HTML]{f3f3f3}1.69 &\cellcolor[HTML]{f3f3f3}0.581 &\cellcolor[HTML]{f3f3f3}\ul{0.233} &\cellcolor[HTML]{f3f3f3}0.65 &\cellcolor[HTML]{f3f3f3}\ul{83.56} &\cellcolor[HTML]{f3f3f3}1.08 &\cellcolor[HTML]{f3f3f3}0.63 &\cellcolor[HTML]{f3f3f3}0.102 &\cellcolor[HTML]{f3f3f3}\ul{0.049} &\cellcolor[HTML]{f3f3f3}0.63 &\cellcolor[HTML]{f3f3f3}96.59 \\ \midrule
  \textbf{DE} &\multirow{3}{*}{90} &2.42 &2.01 &\ul{0.576} &\ul{0.233} &1.79 &83.51 &1.31 &\ul{0.30} &\ul{0.100} &\ul{0.049} &\ul{0.25} &96.66 \\
  \textbf{TDE} & &\cellcolor[HTML]{f3f3f3}\ul{1.39} &\cellcolor[HTML]{f3f3f3}1.58 &\cellcolor[HTML]{f3f3f3}0.579 &\cellcolor[HTML]{f3f3f3}\ul{0.233} &\cellcolor[HTML]{f3f3f3}0.64 &\cellcolor[HTML]{f3f3f3}\ul{83.52} &\cellcolor[HTML]{f3f3f3}\ul{1.11} &\cellcolor[HTML]{f3f3f3}0.55 &\cellcolor[HTML]{f3f3f3}0.102 &\cellcolor[HTML]{f3f3f3}\ul{0.049} &\cellcolor[HTML]{f3f3f3}0.54 &\cellcolor[HTML]{f3f3f3}\ul{96.67} \\
  \textbf{aTDE} & &\cellcolor[HTML]{f3f3f3}1.44 &\cellcolor[HTML]{f3f3f3}\ul{1.57} &\cellcolor[HTML]{f3f3f3}0.579 &\cellcolor[HTML]{f3f3f3}\ul{0.233} &\cellcolor[HTML]{f3f3f3}\ul{0.63} &\cellcolor[HTML]{f3f3f3}83.51 &\cellcolor[HTML]{f3f3f3}\ul{1.11} &\cellcolor[HTML]{f3f3f3}0.56 &\cellcolor[HTML]{f3f3f3}0.102 &\cellcolor[HTML]{f3f3f3}\ul{0.049} &\cellcolor[HTML]{f3f3f3}0.55 &\cellcolor[HTML]{f3f3f3}96.66 \\ \midrule
  \textbf{DE} &\multirow{3}{*}{100} &2.47 &2.00 &\ul{0.576} &\ul{0.233} &1.81 &\ul{83.55} &1.32 &\ul{0.35} &\ul{0.100} &\ul{0.049} &\ul{0.27} &96.67 \\
  \textbf{TDE} & &\cellcolor[HTML]{f3f3f3}1.44 &\cellcolor[HTML]{f3f3f3}\ul{1.61} &\cellcolor[HTML]{f3f3f3}0.578 &\cellcolor[HTML]{f3f3f3}\ul{0.233} &\cellcolor[HTML]{f3f3f3}\ul{0.68} &\cellcolor[HTML]{f3f3f3}83.54 &\cellcolor[HTML]{f3f3f3}\ul{1.08} &\cellcolor[HTML]{f3f3f3}0.52 &\cellcolor[HTML]{f3f3f3}0.102 &\cellcolor[HTML]{f3f3f3}\ul{0.049} &\cellcolor[HTML]{f3f3f3}0.51 &\cellcolor[HTML]{f3f3f3}\ul{96.68} \\
  \textbf{aTDE} & &\cellcolor[HTML]{f3f3f3}\ul{1.42} &\cellcolor[HTML]{f3f3f3}1.62 &\cellcolor[HTML]{f3f3f3}0.578 &\cellcolor[HTML]{f3f3f3}\ul{0.233} &\cellcolor[HTML]{f3f3f3}0.69 &\cellcolor[HTML]{f3f3f3}\ul{83.55} &\cellcolor[HTML]{f3f3f3}\ul{1.09} &\cellcolor[HTML]{f3f3f3}0.53 &\cellcolor[HTML]{f3f3f3}0.102 &\cellcolor[HTML]{f3f3f3}\ul{0.049} &\cellcolor[HTML]{f3f3f3}0.52 &\cellcolor[HTML]{f3f3f3}96.67 \\  
  \bottomrule
  \end{tabular}}
\end{table}
  
% ==================== ==================== WideResNet28x10 ==================== ====================
\begin{table}[!htp]\centering
  \caption{Comparison between DE/TDE/aTDE on WideResNet28x10. The best results are highlighted.}\label{tab:supp-truth3}
  \scriptsize
  \resizebox{\textwidth}{!}{%
  \begin{tabular}{rc|rrrrrr|rrrrrrr}\toprule
  \textbf{} &\textbf{} &\multicolumn{6}{c}{\textbf{CIFAR100}} &\multicolumn{6}{c}{\textbf{CIFAR10}} \\\cmidrule{3-14}
  \textbf{} &\textbf{\#sources} &\textbf{ECE\textsuperscript{\emph{KDE}}$\downarrow$} &\textbf{ECE$\downarrow$} &\textbf{NLL$\downarrow$} &\textbf{MSE$\downarrow$} &\textbf{KS$\downarrow$} &\textbf{ACC$\uparrow$} &\textbf{ECE\textsuperscript{\emph{KDE}}$\downarrow$} &\textbf{ECE$\downarrow$} &\textbf{NLL$\downarrow$} &\textbf{MSE$\downarrow$} &\textbf{KS$\downarrow$} &\textbf{ACC$\uparrow$} \\\midrule
  \textbf{DE} &\multirow{3}{*}{10} &6.00 &5.41 &\ul{0.629} &\ul{0.235} &5.22 &83.97 &\ul{1.04} &\ul{0.24} &\ul{0.094} &\ul{0.045} &\ul{0.21} &\ul{96.93} \\
  \textbf{TDE} & &\cellcolor[HTML]{f3f3f3}4.73 &\cellcolor[HTML]{f3f3f3}4.69 &\cellcolor[HTML]{f3f3f3}0.641 &\cellcolor[HTML]{f3f3f3}0.237 &\cellcolor[HTML]{f3f3f3}4.01 &\cellcolor[HTML]{f3f3f3}\ul{84.10} &\cellcolor[HTML]{f3f3f3}1.28 &\cellcolor[HTML]{f3f3f3}1.06 &\cellcolor[HTML]{f3f3f3}0.099 &\cellcolor[HTML]{f3f3f3}0.047 &\cellcolor[HTML]{f3f3f3}1.05 &\cellcolor[HTML]{f3f3f3}\ul{96.93} \\
  \textbf{aTDE} & &\cellcolor[HTML]{f3f3f3}\ul{4.61} &\cellcolor[HTML]{f3f3f3}\ul{4.59} &\cellcolor[HTML]{f3f3f3}0.641 &\cellcolor[HTML]{f3f3f3}0.237 &\cellcolor[HTML]{f3f3f3}\ul{3.91} &\cellcolor[HTML]{f3f3f3}83.97 &\cellcolor[HTML]{f3f3f3}1.28 &\cellcolor[HTML]{f3f3f3}1.06 &\cellcolor[HTML]{f3f3f3}0.099 &\cellcolor[HTML]{f3f3f3}0.047 &\cellcolor[HTML]{f3f3f3}1.04 &\cellcolor[HTML]{f3f3f3}\ul{96.93} \\ \midrule
  \textbf{DE} &\multirow{3}{*}{20} &6.27 &5.70 &\ul{0.621} &\ul{0.233} &5.49 &84.21 &\ul{0.96} &\ul{0.23} &\ul{0.092} &\ul{0.045} &\ul{0.11} &\ul{97.02} \\
  \textbf{TDE} & &\cellcolor[HTML]{f3f3f3}5.05 &\cellcolor[HTML]{f3f3f3}4.62 &\cellcolor[HTML]{f3f3f3}0.628 &\cellcolor[HTML]{f3f3f3}0.234 &\cellcolor[HTML]{f3f3f3}4.25 &\cellcolor[HTML]{f3f3f3}\ul{84.23} &\cellcolor[HTML]{f3f3f3}1.19 &\cellcolor[HTML]{f3f3f3}0.80 &\cellcolor[HTML]{f3f3f3}0.096 &\cellcolor[HTML]{f3f3f3}0.046 &\cellcolor[HTML]{f3f3f3}0.78 &\cellcolor[HTML]{f3f3f3}97.00 \\
  \textbf{aTDE} & &\cellcolor[HTML]{f3f3f3}\ul{5.04} &\cellcolor[HTML]{f3f3f3}\ul{4.61} &\cellcolor[HTML]{f3f3f3}0.628 &\cellcolor[HTML]{f3f3f3}0.234 &\cellcolor[HTML]{f3f3f3}\ul{4.24} &\cellcolor[HTML]{f3f3f3}84.21 &\cellcolor[HTML]{f3f3f3}1.18 &\cellcolor[HTML]{f3f3f3}0.77 &\cellcolor[HTML]{f3f3f3}0.096 &\cellcolor[HTML]{f3f3f3}0.046 &\cellcolor[HTML]{f3f3f3}0.76 &\cellcolor[HTML]{f3f3f3}\ul{97.02} \\ \midrule
  \textbf{DE} &\multirow{3}{*}{30} &6.28 &5.73 &\ul{0.617} &\ul{0.232} &5.49 &84.22 &\ul{1.07} &\ul{0.33} &\ul{0.091} &\ul{0.045} &\ul{0.24} &\ul{97.15} \\
  \textbf{TDE} & &\cellcolor[HTML]{f3f3f3}5.20 &\cellcolor[HTML]{f3f3f3}4.69 &\cellcolor[HTML]{f3f3f3}0.623 &\cellcolor[HTML]{f3f3f3}0.233 &\cellcolor[HTML]{f3f3f3}4.35 &\cellcolor[HTML]{f3f3f3}\ul{84.26} &\cellcolor[HTML]{f3f3f3}1.15 &\cellcolor[HTML]{f3f3f3}0.60 &\cellcolor[HTML]{f3f3f3}0.094 &\cellcolor[HTML]{f3f3f3}\ul{0.045} &\cellcolor[HTML]{f3f3f3}0.57 &\cellcolor[HTML]{f3f3f3}97.12 \\
  \textbf{aTDE} & &\cellcolor[HTML]{f3f3f3}\ul{5.16} &\cellcolor[HTML]{f3f3f3}\ul{4.66} &\cellcolor[HTML]{f3f3f3}0.623 &\cellcolor[HTML]{f3f3f3}0.233 &\cellcolor[HTML]{f3f3f3}\ul{4.32} &\cellcolor[HTML]{f3f3f3}84.22 &\cellcolor[HTML]{f3f3f3}1.17 &\cellcolor[HTML]{f3f3f3}0.57 &\cellcolor[HTML]{f3f3f3}0.094 &\cellcolor[HTML]{f3f3f3}\ul{0.045} &\cellcolor[HTML]{f3f3f3}0.54 &\cellcolor[HTML]{f3f3f3}\ul{97.15} \\ \midrule
  \textbf{DE} &\multirow{3}{*}{40} &6.39 &5.75 &\ul{0.616} &\ul{0.232} &5.58 &84.28 &\ul{1.09} &\ul{0.28} &\ul{0.091} &\ul{0.044} &\ul{0.21} &\ul{97.14} \\
  \textbf{TDE} & &\cellcolor[HTML]{f3f3f3}5.39 &\cellcolor[HTML]{f3f3f3}4.86 &\cellcolor[HTML]{f3f3f3}0.622 &\cellcolor[HTML]{f3f3f3}\ul{0.232} &\cellcolor[HTML]{f3f3f3}4.54 &\cellcolor[HTML]{f3f3f3}\ul{84.31} &\cellcolor[HTML]{f3f3f3}1.25 &\cellcolor[HTML]{f3f3f3}0.53 &\cellcolor[HTML]{f3f3f3}0.093 &\cellcolor[HTML]{f3f3f3}0.045 &\cellcolor[HTML]{f3f3f3}0.51 &\cellcolor[HTML]{f3f3f3}\ul{97.14} \\
  \textbf{aTDE} & &\cellcolor[HTML]{f3f3f3}\ul{5.36} &\cellcolor[HTML]{f3f3f3}\ul{4.83} &\cellcolor[HTML]{f3f3f3}0.622 &\cellcolor[HTML]{f3f3f3}\ul{0.232} &\cellcolor[HTML]{f3f3f3}\ul{4.51} &\cellcolor[HTML]{f3f3f3}84.28 &\cellcolor[HTML]{f3f3f3}1.27 &\cellcolor[HTML]{f3f3f3}0.53 &\cellcolor[HTML]{f3f3f3}0.093 &\cellcolor[HTML]{f3f3f3}0.045 &\cellcolor[HTML]{f3f3f3}0.51 &\cellcolor[HTML]{f3f3f3}\ul{97.14} \\ \midrule
  \textbf{DE} &\multirow{3}{*}{50} &6.45 &5.86 &\ul{0.615} &\ul{0.232} &5.64 &\ul{84.38} &\ul{1.07} &\ul{0.35} &\ul{0.091} &\ul{0.044} &\ul{0.24} &\ul{97.17} \\
  \textbf{TDE} & &\cellcolor[HTML]{f3f3f3}\ul{5.48} &\cellcolor[HTML]{f3f3f3}\ul{5.04} &\cellcolor[HTML]{f3f3f3}0.621 &\cellcolor[HTML]{f3f3f3}\ul{0.232} &\cellcolor[HTML]{f3f3f3}\ul{4.62} &\cellcolor[HTML]{f3f3f3}84.37 &\cellcolor[HTML]{f3f3f3}1.15 &\cellcolor[HTML]{f3f3f3}0.49 &\cellcolor[HTML]{f3f3f3}0.093 &\cellcolor[HTML]{f3f3f3}0.045 &\cellcolor[HTML]{f3f3f3}0.47 &\cellcolor[HTML]{f3f3f3}97.15 \\
  \textbf{aTDE} & &\cellcolor[HTML]{f3f3f3}5.50 &\cellcolor[HTML]{f3f3f3}5.05 &\cellcolor[HTML]{f3f3f3}0.621 &\cellcolor[HTML]{f3f3f3}\ul{0.232} &\cellcolor[HTML]{f3f3f3}4.63 &\cellcolor[HTML]{f3f3f3}\ul{84.38} &\cellcolor[HTML]{f3f3f3}1.14 &\cellcolor[HTML]{f3f3f3}0.47 &\cellcolor[HTML]{f3f3f3}0.093 &\cellcolor[HTML]{f3f3f3}0.045 &\cellcolor[HTML]{f3f3f3}0.45 &\cellcolor[HTML]{f3f3f3}\ul{97.17} \\ \midrule
  \textbf{DE} &\multirow{3}{*}{60} &6.39 &5.72 &\ul{0.614} &\ul{0.232} &5.57 &84.30 &\ul{1.05} &\ul{0.34} &\ul{0.091} &\ul{0.044} &\ul{0.21} &\ul{97.14} \\
  \textbf{TDE} & &\cellcolor[HTML]{f3f3f3}5.53 &\cellcolor[HTML]{f3f3f3}5.00 &\cellcolor[HTML]{f3f3f3}0.619 &\cellcolor[HTML]{f3f3f3}\ul{0.232} &\cellcolor[HTML]{f3f3f3}4.66 &\cellcolor[HTML]{f3f3f3}\ul{84.36} &\cellcolor[HTML]{f3f3f3}1.14 &\cellcolor[HTML]{f3f3f3}0.49 &\cellcolor[HTML]{f3f3f3}0.093 &\cellcolor[HTML]{f3f3f3}0.045 &\cellcolor[HTML]{f3f3f3}0.47 &\cellcolor[HTML]{f3f3f3}97.13 \\
  \textbf{aTDE} & &\cellcolor[HTML]{f3f3f3}\ul{5.47} &\cellcolor[HTML]{f3f3f3}\ul{4.94} &\cellcolor[HTML]{f3f3f3}0.619 &\cellcolor[HTML]{f3f3f3}\ul{0.232} &\cellcolor[HTML]{f3f3f3}\ul{4.60} &\cellcolor[HTML]{f3f3f3}84.30 &\cellcolor[HTML]{f3f3f3}1.13 &\cellcolor[HTML]{f3f3f3}0.48 &\cellcolor[HTML]{f3f3f3}0.093 &\cellcolor[HTML]{f3f3f3}0.045 &\cellcolor[HTML]{f3f3f3}0.46 &\cellcolor[HTML]{f3f3f3}\ul{97.14} \\ \midrule
  \textbf{DE} &\multirow{3}{*}{70} &6.42 &5.75 &\ul{0.614} &\ul{0.232} &5.59 &84.31 &\ul{1.10} &\ul{0.32} &\ul{0.090} &\ul{0.044} &\ul{0.23} &\ul{97.15} \\
  \textbf{TDE} & &\cellcolor[HTML]{f3f3f3}5.55 &\cellcolor[HTML]{f3f3f3}5.02 &\cellcolor[HTML]{f3f3f3}0.619 &\cellcolor[HTML]{f3f3f3}\ul{0.232} &\cellcolor[HTML]{f3f3f3}4.70 &\cellcolor[HTML]{f3f3f3}\ul{84.34} &\cellcolor[HTML]{f3f3f3}\ul{1.10} &\cellcolor[HTML]{f3f3f3}0.46 &\cellcolor[HTML]{f3f3f3}0.092 &\cellcolor[HTML]{f3f3f3}0.045 &\cellcolor[HTML]{f3f3f3}0.44 &\cellcolor[HTML]{f3f3f3}97.14 \\
  \textbf{aTDE} & &\cellcolor[HTML]{f3f3f3}\ul{5.52} &\cellcolor[HTML]{f3f3f3}\ul{4.99} &\cellcolor[HTML]{f3f3f3}0.619 &\cellcolor[HTML]{f3f3f3}\ul{0.232} &\cellcolor[HTML]{f3f3f3}\ul{4.67} &\cellcolor[HTML]{f3f3f3}84.31 &\cellcolor[HTML]{f3f3f3}1.11 &\cellcolor[HTML]{f3f3f3}0.45 &\cellcolor[HTML]{f3f3f3}0.092 &\cellcolor[HTML]{f3f3f3}0.045 &\cellcolor[HTML]{f3f3f3}0.43 &\cellcolor[HTML]{f3f3f3}\ul{97.15} \\ \midrule
  \textbf{DE} &\multirow{3}{*}{80} &6.44 &5.80 &\ul{0.615} &\ul{0.232} &5.62 &\ul{84.30} &1.14 &\ul{0.33} &\ul{0.090} &\ul{0.044} &\ul{0.26} &\ul{97.16} \\
  \textbf{TDE} & &\cellcolor[HTML]{f3f3f3}\ul{5.54} &\cellcolor[HTML]{f3f3f3}\ul{5.00} &\cellcolor[HTML]{f3f3f3}0.619 &\cellcolor[HTML]{f3f3f3}\ul{0.232} &\cellcolor[HTML]{f3f3f3}\ul{4.70} &\cellcolor[HTML]{f3f3f3}84.27 &\cellcolor[HTML]{f3f3f3}\ul{1.10} &\cellcolor[HTML]{f3f3f3}0.45 &\cellcolor[HTML]{f3f3f3}0.092 &\cellcolor[HTML]{f3f3f3}\ul{0.044} &\cellcolor[HTML]{f3f3f3}0.41 &\cellcolor[HTML]{f3f3f3}97.15 \\
  \textbf{aTDE} & &\cellcolor[HTML]{f3f3f3}5.57 &\cellcolor[HTML]{f3f3f3}5.03 &\cellcolor[HTML]{f3f3f3}0.619 &\cellcolor[HTML]{f3f3f3}\ul{0.232} &\cellcolor[HTML]{f3f3f3}4.74 &\cellcolor[HTML]{f3f3f3}\ul{84.30} &\cellcolor[HTML]{f3f3f3}1.11 &\cellcolor[HTML]{f3f3f3}0.44 &\cellcolor[HTML]{f3f3f3}0.092 &\cellcolor[HTML]{f3f3f3}\ul{0.044} &\cellcolor[HTML]{f3f3f3}0.40 &\cellcolor[HTML]{f3f3f3}\ul{97.16} \\ \midrule
  \textbf{DE} &\multirow{3}{*}{90} &6.43 &5.83 &\ul{0.614} &\ul{0.232} &5.63 &84.30 &1.20 &0.42 &\ul{0.090} &\ul{0.044} &\ul{0.29} &\ul{97.19} \\
  \textbf{TDE} & &\cellcolor[HTML]{f3f3f3}5.59 &\cellcolor[HTML]{f3f3f3}5.11 &\cellcolor[HTML]{f3f3f3}0.619 &\cellcolor[HTML]{f3f3f3}\ul{0.232} &\cellcolor[HTML]{f3f3f3}4.75 &\cellcolor[HTML]{f3f3f3}\ul{84.31} &\cellcolor[HTML]{f3f3f3}\ul{1.12} &\cellcolor[HTML]{f3f3f3}\ul{0.39} &\cellcolor[HTML]{f3f3f3}0.092 &\cellcolor[HTML]{f3f3f3}\ul{0.044} &\cellcolor[HTML]{f3f3f3}0.36 &\cellcolor[HTML]{f3f3f3}\ul{97.19} \\
  \textbf{aTDE} & &\cellcolor[HTML]{f3f3f3}\ul{5.58} &\cellcolor[HTML]{f3f3f3}\ul{5.10} &\cellcolor[HTML]{f3f3f3}0.619 &\cellcolor[HTML]{f3f3f3}\ul{0.232} &\cellcolor[HTML]{f3f3f3}\ul{4.74} &\cellcolor[HTML]{f3f3f3}84.30 &\cellcolor[HTML]{f3f3f3}\ul{1.12} &\cellcolor[HTML]{f3f3f3}\ul{0.39} &\cellcolor[HTML]{f3f3f3}0.092 &\cellcolor[HTML]{f3f3f3}\ul{0.044} &\cellcolor[HTML]{f3f3f3}0.36 &\cellcolor[HTML]{f3f3f3}\ul{97.19} \\ \midrule
  \textbf{DE} &\multirow{3}{*}{100} &6.41 &5.79 &\ul{0.614} &\ul{0.231} &5.61 &84.28 &1.20 &0.41 &\ul{0.090} &\ul{0.044} &\ul{0.30} &\ul{97.20} \\
  \textbf{TDE} & &\cellcolor[HTML]{f3f3f3}5.58 &\cellcolor[HTML]{f3f3f3}5.08 &\cellcolor[HTML]{f3f3f3}0.618 &\cellcolor[HTML]{f3f3f3}0.232 &\cellcolor[HTML]{f3f3f3}4.74 &\cellcolor[HTML]{f3f3f3}\ul{84.29} &\cellcolor[HTML]{f3f3f3}\ul{1.11} &\cellcolor[HTML]{f3f3f3}\ul{0.40} &\cellcolor[HTML]{f3f3f3}0.092 &\cellcolor[HTML]{f3f3f3}\ul{0.044} &\cellcolor[HTML]{f3f3f3}0.36 &\cellcolor[HTML]{f3f3f3}97.17 \\
  \textbf{aTDE} & &\cellcolor[HTML]{f3f3f3}\ul{5.57} &\cellcolor[HTML]{f3f3f3}\ul{5.07} &\cellcolor[HTML]{f3f3f3}0.618 &\cellcolor[HTML]{f3f3f3}0.232 &\cellcolor[HTML]{f3f3f3}\ul{4.73} &\cellcolor[HTML]{f3f3f3}84.28 &\cellcolor[HTML]{f3f3f3}\ul{1.13} &\cellcolor[HTML]{f3f3f3}0.42 &\cellcolor[HTML]{f3f3f3}0.092 &\cellcolor[HTML]{f3f3f3}\ul{0.044} &\cellcolor[HTML]{f3f3f3}0.33 &\cellcolor[HTML]{f3f3f3}\ul{97.20} \\  
  \bottomrule
  \end{tabular}}
\end{table}

% ==================== ==================== ImageNet ==================== ====================
\begin{table}[!htp]\centering
  \caption{Comparison between DE/TDE/aTDE on ImageNet. The best results are highlighted.}\label{tab:supp-truth4}
  \scriptsize
  \resizebox{\textwidth}{!}{%
  \begin{tabular}{rc|rrrrrr|rrrrrrr}\toprule
  \textbf{} &\textbf{} &\multicolumn{6}{c}{\textbf{ImageNet}} &\multicolumn{6}{c}{\textbf{ImageNet (Snapshot Ensemble)}} \\\cmidrule{3-14}
  \textbf{} &\textbf{\#sources} &\textbf{ECE\textsuperscript{\emph{KDE}}$\downarrow$} &\textbf{ECE$\downarrow$} &\textbf{NLL$\downarrow$} &\textbf{MSE$\downarrow$} &\textbf{KS$\downarrow$} &\textbf{ACC$\uparrow$} &\textbf{ECE\textsuperscript{\emph{KDE}}$\downarrow$} &\textbf{ECE$\downarrow$} &\textbf{NLL$\downarrow$} &\textbf{MSE$\downarrow$} &\textbf{KS$\downarrow$} &\textbf{ACC$\uparrow$} \\\midrule
  \textbf{DE} &\multirow{3}{*}{5} &2.32 &2.45 &\ul{0.831} &\ul{0.302} &2.04 &\ul{78.65} &2.00 &\ul{2.18} &\ul{0.866} &\ul{0.313} &\ul{2.18} &\ul{77.79} \\
  \textbf{TDE} & &\cellcolor[HTML]{f3f3f3}\ul{1.99} &\cellcolor[HTML]{f3f3f3}\ul{2.43} &\cellcolor[HTML]{f3f3f3}0.838 &\cellcolor[HTML]{f3f3f3}0.304 &\cellcolor[HTML]{f3f3f3}\ul{0.90} &\cellcolor[HTML]{f3f3f3}78.38 &\cellcolor[HTML]{f3f3f3}\ul{1.99} &\cellcolor[HTML]{f3f3f3}2.46 &\cellcolor[HTML]{f3f3f3}0.874 &\cellcolor[HTML]{f3f3f3}0.315 &\cellcolor[HTML]{f3f3f3}2.46 &\cellcolor[HTML]{f3f3f3}77.61 \\
  \textbf{aTDE} & &\cellcolor[HTML]{f3f3f3}2.33 &\cellcolor[HTML]{f3f3f3}2.78 &\cellcolor[HTML]{f3f3f3}0.837 &\cellcolor[HTML]{f3f3f3}0.304 &\cellcolor[HTML]{f3f3f3}1.26 &\cellcolor[HTML]{f3f3f3}\ul{78.65} &\cellcolor[HTML]{f3f3f3}2.21 &\cellcolor[HTML]{f3f3f3}2.69 &\cellcolor[HTML]{f3f3f3}0.873 &\cellcolor[HTML]{f3f3f3}0.315 &\cellcolor[HTML]{f3f3f3}2.69 &\cellcolor[HTML]{f3f3f3}\ul{77.79} \\ \midrule
  \textbf{DE} &\multirow{3}{*}{10} &2.94 &2.94 &\ul{0.808} &\ul{0.297} &2.60 &\ul{79.15} &2.12 &2.32 &\ul{0.843} &\ul{0.307} &2.32 &\ul{78.25} \\
  \textbf{TDE} & &\cellcolor[HTML]{f3f3f3}\ul{1.96} &\cellcolor[HTML]{f3f3f3}\ul{2.31} &\cellcolor[HTML]{f3f3f3}0.809 &\cellcolor[HTML]{f3f3f3}\ul{0.297} &\cellcolor[HTML]{f3f3f3}\ul{1.41} &\cellcolor[HTML]{f3f3f3}79.02 &\cellcolor[HTML]{f3f3f3}\ul{1.78} &\cellcolor[HTML]{f3f3f3}\ul{2.17} &\cellcolor[HTML]{f3f3f3}0.846 &\cellcolor[HTML]{f3f3f3}\ul{0.307} &\cellcolor[HTML]{f3f3f3}\ul{2.17} &\cellcolor[HTML]{f3f3f3}78.17 \\
  \textbf{aTDE} & &\cellcolor[HTML]{f3f3f3}2.11 &\cellcolor[HTML]{f3f3f3}2.46 &\cellcolor[HTML]{f3f3f3}0.809 &\cellcolor[HTML]{f3f3f3}\ul{0.297} &\cellcolor[HTML]{f3f3f3}1.56 &\cellcolor[HTML]{f3f3f3}\ul{79.15} &\cellcolor[HTML]{f3f3f3}1.88 &\cellcolor[HTML]{f3f3f3}2.26 &\cellcolor[HTML]{f3f3f3}0.846 &\cellcolor[HTML]{f3f3f3}\ul{0.307} &\cellcolor[HTML]{f3f3f3}2.26 &\cellcolor[HTML]{f3f3f3}\ul{78.25} \\ \midrule
  \textbf{DE} &\multirow{3}{*}{15} &2.96 &2.92 &0.802 &\ul{0.295} &2.61 &79.15 &1.98 &2.20 &\ul{0.837} &\ul{0.304} &2.20 &\ul{78.37} \\
  \textbf{TDE} & &\cellcolor[HTML]{f3f3f3}2.04 &\cellcolor[HTML]{f3f3f3}2.32 &\cellcolor[HTML]{f3f3f3}\ul{0.801} &\cellcolor[HTML]{f3f3f3}\ul{0.295} &\cellcolor[HTML]{f3f3f3}1.60 &\cellcolor[HTML]{f3f3f3}\ul{79.17} &\cellcolor[HTML]{f3f3f3}\ul{1.74} &\cellcolor[HTML]{f3f3f3}\ul{2.06} &\cellcolor[HTML]{f3f3f3}0.839 &\cellcolor[HTML]{f3f3f3}0.305 &\cellcolor[HTML]{f3f3f3}\ul{2.06} &\cellcolor[HTML]{f3f3f3}78.32 \\
  \textbf{aTDE} & &\cellcolor[HTML]{f3f3f3}\ul{2.03} &\cellcolor[HTML]{f3f3f3}\ul{2.31} &\cellcolor[HTML]{f3f3f3}\ul{0.801} &\cellcolor[HTML]{f3f3f3}\ul{0.295} &\cellcolor[HTML]{f3f3f3}\ul{1.59} &\cellcolor[HTML]{f3f3f3}79.15 &\cellcolor[HTML]{f3f3f3}1.79 &\cellcolor[HTML]{f3f3f3}2.12 &\cellcolor[HTML]{f3f3f3}0.839 &\cellcolor[HTML]{f3f3f3}0.305 &\cellcolor[HTML]{f3f3f3}2.12 &\cellcolor[HTML]{f3f3f3}\ul{78.37} \\ \midrule
  \textbf{DE} &\multirow{3}{*}{20} &3.10 &3.07 &0.799 &0.295 &2.76 &\ul{79.26} &1.89 &2.14 &\ul{0.834} &\ul{0.304} &2.14 &\ul{78.42} \\
  \textbf{TDE} & &\cellcolor[HTML]{f3f3f3}\ul{2.14} &\cellcolor[HTML]{f3f3f3}\ul{2.36} &\cellcolor[HTML]{f3f3f3}\ul{0.798} &\cellcolor[HTML]{f3f3f3}\ul{0.294} &\cellcolor[HTML]{f3f3f3}\ul{1.75} &\cellcolor[HTML]{f3f3f3}79.21 &\cellcolor[HTML]{f3f3f3}\ul{1.65} &\cellcolor[HTML]{f3f3f3}\ul{2.01} &\cellcolor[HTML]{f3f3f3}0.837 &\cellcolor[HTML]{f3f3f3}\ul{0.304} &\cellcolor[HTML]{f3f3f3}\ul{2.01} &\cellcolor[HTML]{f3f3f3}78.35 \\
  \textbf{aTDE} & &\cellcolor[HTML]{f3f3f3}2.20 &\cellcolor[HTML]{f3f3f3}2.42 &\cellcolor[HTML]{f3f3f3}\ul{0.798} &\cellcolor[HTML]{f3f3f3}\ul{0.294} &\cellcolor[HTML]{f3f3f3}1.81 &\cellcolor[HTML]{f3f3f3}\ul{79.26} &\cellcolor[HTML]{f3f3f3}1.73 &\cellcolor[HTML]{f3f3f3}2.08 &\cellcolor[HTML]{f3f3f3}0.837 &\cellcolor[HTML]{f3f3f3}\ul{0.304} &\cellcolor[HTML]{f3f3f3}2.08 &\cellcolor[HTML]{f3f3f3}\ul{78.42} \\ \midrule
  \textbf{DE} &\multirow{3}{*}{25} &3.11 &3.12 &0.797 &0.295 &2.79 &\ul{79.25} &1.86 &2.12 &\ul{0.833} &\ul{0.303} &2.12 &\ul{78.46} \\
  \textbf{TDE} & &\cellcolor[HTML]{f3f3f3}\ul{2.17} &\cellcolor[HTML]{f3f3f3}\ul{2.42} &\cellcolor[HTML]{f3f3f3}\ul{0.796} &\cellcolor[HTML]{f3f3f3}\ul{0.294} &\cellcolor[HTML]{f3f3f3}\ul{1.81} &\cellcolor[HTML]{f3f3f3}79.22 &\cellcolor[HTML]{f3f3f3}\ul{1.69} &\cellcolor[HTML]{f3f3f3}\ul{2.09} &\cellcolor[HTML]{f3f3f3}0.836 &\cellcolor[HTML]{f3f3f3}\ul{0.303} &\cellcolor[HTML]{f3f3f3}\ul{2.09} &\cellcolor[HTML]{f3f3f3}78.45 \\
  \textbf{aTDE} & &\cellcolor[HTML]{f3f3f3}2.20 &\cellcolor[HTML]{f3f3f3}2.45 &\cellcolor[HTML]{f3f3f3}\ul{0.796} &\cellcolor[HTML]{f3f3f3}\ul{0.294} &\cellcolor[HTML]{f3f3f3}1.84 &\cellcolor[HTML]{f3f3f3}\ul{79.25} &\cellcolor[HTML]{f3f3f3}1.71 &\cellcolor[HTML]{f3f3f3}2.11 &\cellcolor[HTML]{f3f3f3}0.836 &\cellcolor[HTML]{f3f3f3}\ul{0.303} &\cellcolor[HTML]{f3f3f3}2.11 &\cellcolor[HTML]{f3f3f3}\ul{78.46} \\ \midrule
  \textbf{DE} &\multirow{3}{*}{30} &3.11 &3.09 &0.795 &0.294 &2.78 &\ul{79.27} &1.82 &\ul{2.08} &\ul{0.832} &\ul{0.303} &\ul{2.08} &\ul{78.48} \\
  \textbf{TDE} & &\cellcolor[HTML]{f3f3f3}\ul{2.24} &\cellcolor[HTML]{f3f3f3}\ul{2.47} &\cellcolor[HTML]{f3f3f3}\ul{0.794} &\cellcolor[HTML]{f3f3f3}\ul{0.293} &\cellcolor[HTML]{f3f3f3}\ul{1.88} &\cellcolor[HTML]{f3f3f3}79.26 &\cellcolor[HTML]{f3f3f3}\ul{1.72} &\cellcolor[HTML]{f3f3f3}2.11 &\cellcolor[HTML]{f3f3f3}0.835 &\cellcolor[HTML]{f3f3f3}\ul{0.303} &\cellcolor[HTML]{f3f3f3}2.11 &\cellcolor[HTML]{f3f3f3}78.45 \\
  \textbf{aTDE} & &\cellcolor[HTML]{f3f3f3}\ul{2.24} &\cellcolor[HTML]{f3f3f3}\ul{2.47} &\cellcolor[HTML]{f3f3f3}\ul{0.794} &\cellcolor[HTML]{f3f3f3}\ul{0.293} &\cellcolor[HTML]{f3f3f3}\ul{1.88} &\cellcolor[HTML]{f3f3f3}\ul{79.27} &\cellcolor[HTML]{f3f3f3}1.75 &\cellcolor[HTML]{f3f3f3}2.14 &\cellcolor[HTML]{f3f3f3}0.835 &\cellcolor[HTML]{f3f3f3}\ul{0.303} &\cellcolor[HTML]{f3f3f3}2.14 &\cellcolor[HTML]{f3f3f3}\ul{78.48} \\ \midrule
  \textbf{DE} &\multirow{3}{*}{35} &3.14 &3.10 &0.794 &0.294 &2.80 &\ul{79.27} &1.79 &\ul{2.07} &\ul{0.831} &\ul{0.303} &\ul{2.07} &\ul{78.50} \\
  \textbf{TDE} & &\cellcolor[HTML]{f3f3f3}\ul{2.28} &\cellcolor[HTML]{f3f3f3}\ul{2.49} &\cellcolor[HTML]{f3f3f3}\ul{0.793} &\cellcolor[HTML]{f3f3f3}\ul{0.293} &\cellcolor[HTML]{f3f3f3}\ul{1.92} &\cellcolor[HTML]{f3f3f3}79.26 &\cellcolor[HTML]{f3f3f3}\ul{1.73} &\cellcolor[HTML]{f3f3f3}2.14 &\cellcolor[HTML]{f3f3f3}0.834 &\cellcolor[HTML]{f3f3f3}\ul{0.303} &\cellcolor[HTML]{f3f3f3}2.14 &\cellcolor[HTML]{f3f3f3}\ul{78.50} \\
  \textbf{aTDE} & &\cellcolor[HTML]{f3f3f3}2.29 &\cellcolor[HTML]{f3f3f3}2.50 &\cellcolor[HTML]{f3f3f3}\ul{0.793} &\cellcolor[HTML]{f3f3f3}\ul{0.293} &\cellcolor[HTML]{f3f3f3}1.94 &\cellcolor[HTML]{f3f3f3}\ul{79.27} &\cellcolor[HTML]{f3f3f3}1.74 &\cellcolor[HTML]{f3f3f3}2.15 &\cellcolor[HTML]{f3f3f3}0.834 &\cellcolor[HTML]{f3f3f3}\ul{0.303} &\cellcolor[HTML]{f3f3f3}2.15 &\cellcolor[HTML]{f3f3f3}\ul{78.50} \\ \midrule
  \textbf{DE} &\multirow{3}{*}{40} &3.17 &3.12 &0.794 &0.294 &2.82 &\ul{79.30} &1.77 &\ul{2.06} &\ul{0.830} &\ul{0.302} &\ul{2.06} &\ul{78.51} \\
  \textbf{TDE} & &\cellcolor[HTML]{f3f3f3}\ul{2.31} &\cellcolor[HTML]{f3f3f3}\ul{2.50} &\cellcolor[HTML]{f3f3f3}\ul{0.793} &\cellcolor[HTML]{f3f3f3}\ul{0.293} &\cellcolor[HTML]{f3f3f3}\ul{1.97} &\cellcolor[HTML]{f3f3f3}79.29 &\cellcolor[HTML]{f3f3f3}\ul{1.72} &\cellcolor[HTML]{f3f3f3}2.13 &\cellcolor[HTML]{f3f3f3}0.833 &\cellcolor[HTML]{f3f3f3}\ul{0.302} &\cellcolor[HTML]{f3f3f3}2.13 &\cellcolor[HTML]{f3f3f3}\ul{78.51} \\
  \textbf{aTDE} & &\cellcolor[HTML]{f3f3f3}2.33 &\cellcolor[HTML]{f3f3f3}2.52 &\cellcolor[HTML]{f3f3f3}\ul{0.793} &\cellcolor[HTML]{f3f3f3}\ul{0.293} &\cellcolor[HTML]{f3f3f3}1.98 &\cellcolor[HTML]{f3f3f3}\ul{79.30} &\cellcolor[HTML]{f3f3f3}\ul{1.72} &\cellcolor[HTML]{f3f3f3}2.14 &\cellcolor[HTML]{f3f3f3}0.833 &\cellcolor[HTML]{f3f3f3}\ul{0.302} &\cellcolor[HTML]{f3f3f3}2.14 &\cellcolor[HTML]{f3f3f3}\ul{78.51} \\ \midrule
  \textbf{DE} &\multirow{3}{*}{45} &3.21 &3.14 &0.793 &0.294 &2.86 &\ul{79.34} &1.78 &\ul{2.06} &\ul{0.830} &\ul{0.302} &\ul{2.06} &\ul{78.54} \\
  \textbf{TDE} & &\cellcolor[HTML]{f3f3f3}\ul{2.35} &\cellcolor[HTML]{f3f3f3}\ul{2.54} &\cellcolor[HTML]{f3f3f3}\ul{0.792} &\cellcolor[HTML]{f3f3f3}\ul{0.293} &\cellcolor[HTML]{f3f3f3}\ul{2.01} &\cellcolor[HTML]{f3f3f3}79.31 &\cellcolor[HTML]{f3f3f3}\ul{1.72} &\cellcolor[HTML]{f3f3f3}2.18 &\cellcolor[HTML]{f3f3f3}0.833 &\cellcolor[HTML]{f3f3f3}\ul{0.302} &\cellcolor[HTML]{f3f3f3}2.18 &\cellcolor[HTML]{f3f3f3}78.52 \\
  \textbf{aTDE} & &\cellcolor[HTML]{f3f3f3}2.38 &\cellcolor[HTML]{f3f3f3}2.58 &\cellcolor[HTML]{f3f3f3}\ul{0.792} &\cellcolor[HTML]{f3f3f3}\ul{0.293} &\cellcolor[HTML]{f3f3f3}2.04 &\cellcolor[HTML]{f3f3f3}\ul{79.34} &\cellcolor[HTML]{f3f3f3}1.74 &\cellcolor[HTML]{f3f3f3}2.20 &\cellcolor[HTML]{f3f3f3}0.833 &\cellcolor[HTML]{f3f3f3}\ul{0.302} &\cellcolor[HTML]{f3f3f3}2.20 &\cellcolor[HTML]{f3f3f3}\ul{78.54} \\ \midrule
  \textbf{DE} &\multirow{3}{*}{50} &3.24 &3.17 &0.792 &0.294 &2.89 &\ul{79.37} &1.73 &\ul{2.04} &\ul{0.830} &\ul{0.302} &\ul{2.04} &\ul{78.52} \\
  \textbf{TDE} & &\cellcolor[HTML]{f3f3f3}\ul{2.41} &\cellcolor[HTML]{f3f3f3}\ul{2.58} &\cellcolor[HTML]{f3f3f3}\ul{0.791} &\cellcolor[HTML]{f3f3f3}\ul{0.293} &\cellcolor[HTML]{f3f3f3}\ul{2.07} &\cellcolor[HTML]{f3f3f3}79.35 &\cellcolor[HTML]{f3f3f3}\ul{1.66} &\cellcolor[HTML]{f3f3f3}2.11 &\cellcolor[HTML]{f3f3f3}0.833 &\cellcolor[HTML]{f3f3f3}\ul{0.302} &\cellcolor[HTML]{f3f3f3}2.11 &\cellcolor[HTML]{f3f3f3}78.50 \\
  \textbf{aTDE} & &\cellcolor[HTML]{f3f3f3}2.44 &\cellcolor[HTML]{f3f3f3}2.61 &\cellcolor[HTML]{f3f3f3}\ul{0.791} &\cellcolor[HTML]{f3f3f3}\ul{0.293} &\cellcolor[HTML]{f3f3f3}2.10 &\cellcolor[HTML]{f3f3f3}\ul{79.37} &\cellcolor[HTML]{f3f3f3}1.69 &\cellcolor[HTML]{f3f3f3}2.14 &\cellcolor[HTML]{f3f3f3}0.833 &\cellcolor[HTML]{f3f3f3}\ul{0.302} &\cellcolor[HTML]{f3f3f3}2.14 &\cellcolor[HTML]{f3f3f3}\ul{78.52} \\
  \bottomrule
  \end{tabular}}
\end{table}  

\clearpage
% \subsection{Additional Figures}\label{supp:figs}
% ~~~~~~~~~~~~~~~~~~~~ ~~~~~~~~~~~~~~~~~~~~ ~~~~~~~~~~~~~~~~~~~~ figure: correlations
\begin{figure*}[!ht]
  \centering
  \includegraphics[width=0.99\linewidth]{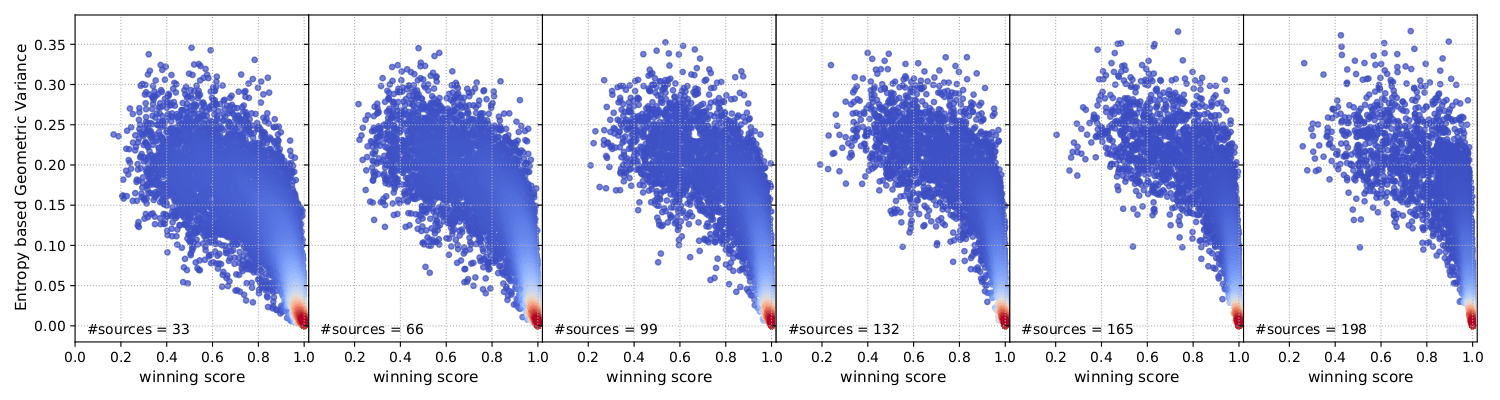}
  \caption{Correlation between Entropy based Geometric Variance (HV) and winning score of DenseNet121 trained on CIFAR10. Winning score stands for the score of predicted label, i.e. the maximum score of the network's output. }\label{supp:cifar10-densnet}
\end{figure*}

\begin{figure*}[!ht]
  \centering
  \includegraphics[width=0.99\linewidth]{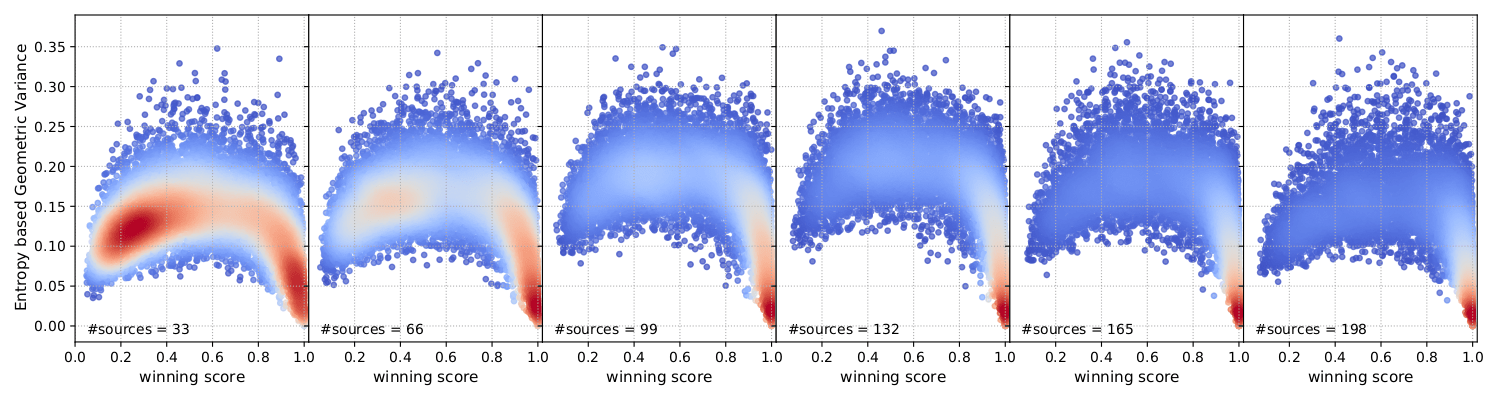}
  \caption{Correlation between Entropy based Geometric Variance (HV) and winning score of ResNet18 trained on CIFAR100.}\label{supp:cifar100-resnet}
\end{figure*}

%\begin{figure*}[!ht]
%  \centering
%  \includegraphics[width=0.99\linewidth]{pic/imagenet_sse-min.png}
%  \caption{Correlation between Entropy based Geometric Variance (HV) and winning score of ResNet50 trained on ImageNet using SE.}\label{supp:imagenet-sse}
%\end{figure*}

\begin{figure*}[!ht]
  \centering
  \includegraphics[width=0.99\linewidth]{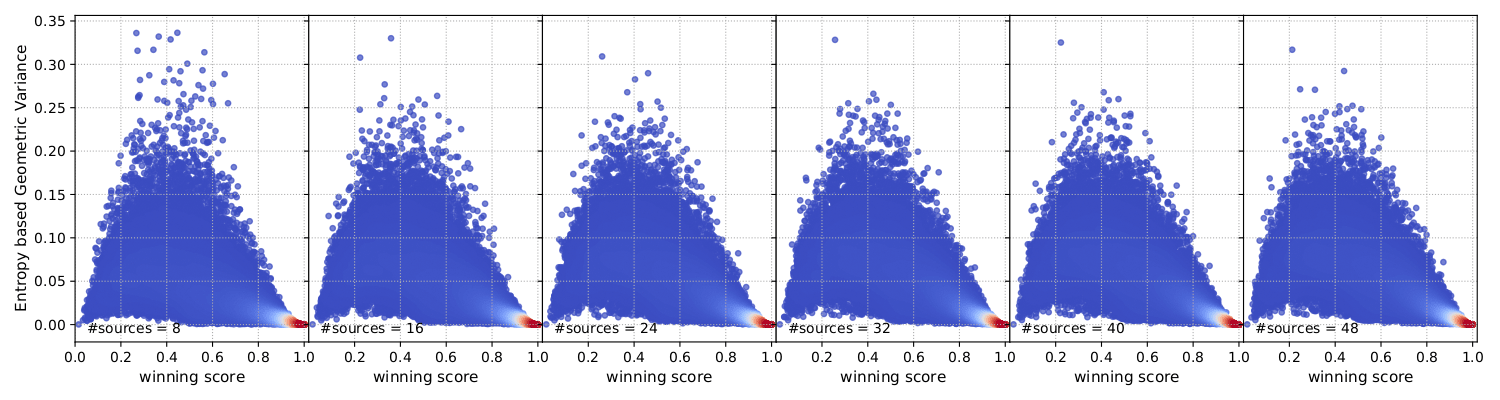}
  \caption{Correlation of Entropy based Geometric Variance (HV) and winning score of ResNet50 trained on ImageNet using DE.}\label{supp:imagenet-dee}
\end{figure*}

% ==================== ==================== ==================== ==================== ====================
%                                             End of Everything                                          %
% ==================== ==================== ==================== ==================== ====================  
\end{document}